\documentclass{bmvc2k}

\usepackage[utf8]{inputenc}
\usepackage{amsmath, amssymb, amsfonts}
\usepackage{bm}
\usepackage{dsfont}
\usepackage{algorithm}
\usepackage{booktabs}
\usepackage{xcolor}
\usepackage{hyperref}
\usepackage{graphicx}
\usepackage{dblfloatfix}
\usepackage{algpseudocode}
\usepackage{float}
\usepackage{enumitem}

\title{Adaptive Tokenisation Via Temporal Redundancy Masking And Latent Inpainting}

\addauthor{Kevin Dave\footnotemark[1]}{kevindave5735@gmail.com}{1}
\addauthor{Sai Aditya Patkuri\footnotemark[1]}{psaiaditya@iisc.ac.in}{1, 2}
\addauthor{Chhaya Kumar Das\footnotemark[1]}{chayakudas.das56@gmail.com}{1}
\addauthor{Gouranga Bala}{gouranga.bala23@gmail.com}{1}
\addauthor{Rajesh Kumar S A}{rajeshkumar.s.a@gmail.com}{1}
\addauthor{R. Venkatesh Babu}{venky@iisc.ac.in}{2}

\footnotetext[1]{Equal contribution.}

\addinstitution{
Phronetic AI
}
\addinstitution{
Vision and AI Lab,\\
Indian Institute of Science,\\
Bengaluru, India
}

\runninghead{Kevin, Sai, Chhaya, Gouranga, Rajesh, Babu}{Adaptive Tokenisation}
\begin{document}

\maketitle
\vspace{-2em}

\begin{abstract}

Adaptive video tokenisation seeks to dynamically allocate token budgets based on the underlying visual complexity of a sequence. Current continuous-regime approaches achieve this via iterative binarised searches or trained neural regressors, while discrete methods often require a full-rate decoder pass to estimate information content. We demonstrate that such computational overheads are not strictly necessary. We show that the latent space of a frozen continuous video tokeniser inherently encodes temporal redundancy that can be exploited directly: spatial positions whose latent representations change minimally between consecutive frames carry near-zero additional information.

We introduce a parameter-free adaptive token allocation mechanism that applies a fixed threshold to per-position temporal-L1 differences, identifying and dropping redundant latent positions. Consequently, the compression rate emerges naturally from the input content rather than being enforced top-down: static scenes get compressed aggressively, while highly dynamic sequences retain more tokens. To reconstruct the dropped positions, we propose the Latent Inpainting Transformer (LIT), a lightweight factorised spatial-temporal attention architecture. The resulting inference pipeline is highly efficient, requiring only a single encoder pass and one LIT forward pass, eliminating the need for auxiliary routing networks. Evaluations across TokenBench and DAVIS, which are the standard benchmarks used by recent tokenisers~\cite{infotok, agarwal2025cosmos}, indicate that our framework yields
meaningful, content-driven token allocation while maintaining competitive reconstruction fidelity, and delivers a $31\times$ inference-time speedup over the continuous adaptive baseline (ElasticTok-CV) and an $\approx2\times$ speedup over the discrete information-theoretic baseline (InfoTok). Project page: \url{https://apatkuri.github.io/adaptive-tokenisation-bmvc-2026/}

\end{abstract}

\section{Introduction}
\label{sec:intro}


Video foundation models are becoming increasingly important in contemporary artificial intelligence (AI) research, driven by the ambition to construct reliable representations of the world \cite{videoworldsimulators2024, assran2024learning}. Crucial to this paradigm are tokenisers, which are responsible for compressing high-dimensional, raw video data $X \in \mathbb{R}^{T \times H \times W \times C}$ into compact, discrete or low-rank representations \cite{yan2021videogpt, yu2024language, nvidia2024cosmos, mondal2020maskaaelatentspaceoptimization}.

Video data is natively characterised by a highly non-uniform spatiotemporal 
information density \cite{yan2025elastictok, agarwal2025cosmos}. While Transformer-based 
architectures \cite{vaswani2017attention, dosovitskiy2020image, arnab2021vivit} have 
emerged as a highly effective paradigm for packing high-dimensional visual information 
into structured sequence representations, a critical architectural bottleneck remains. 
Prevailing spatiotemporal tokenisers compress video sequences using a rigid, fixed 
token allocation budget across all inputs \cite{esser2021taming, yan2021videogpt, 
videoworldsimulators2024, agarwal2025cosmos}, not taking into account the underlying variations in 
dynamic complexity or content redundancy.

\begin{figure*}[t]
    \centering
    \includegraphics[width=\textwidth]{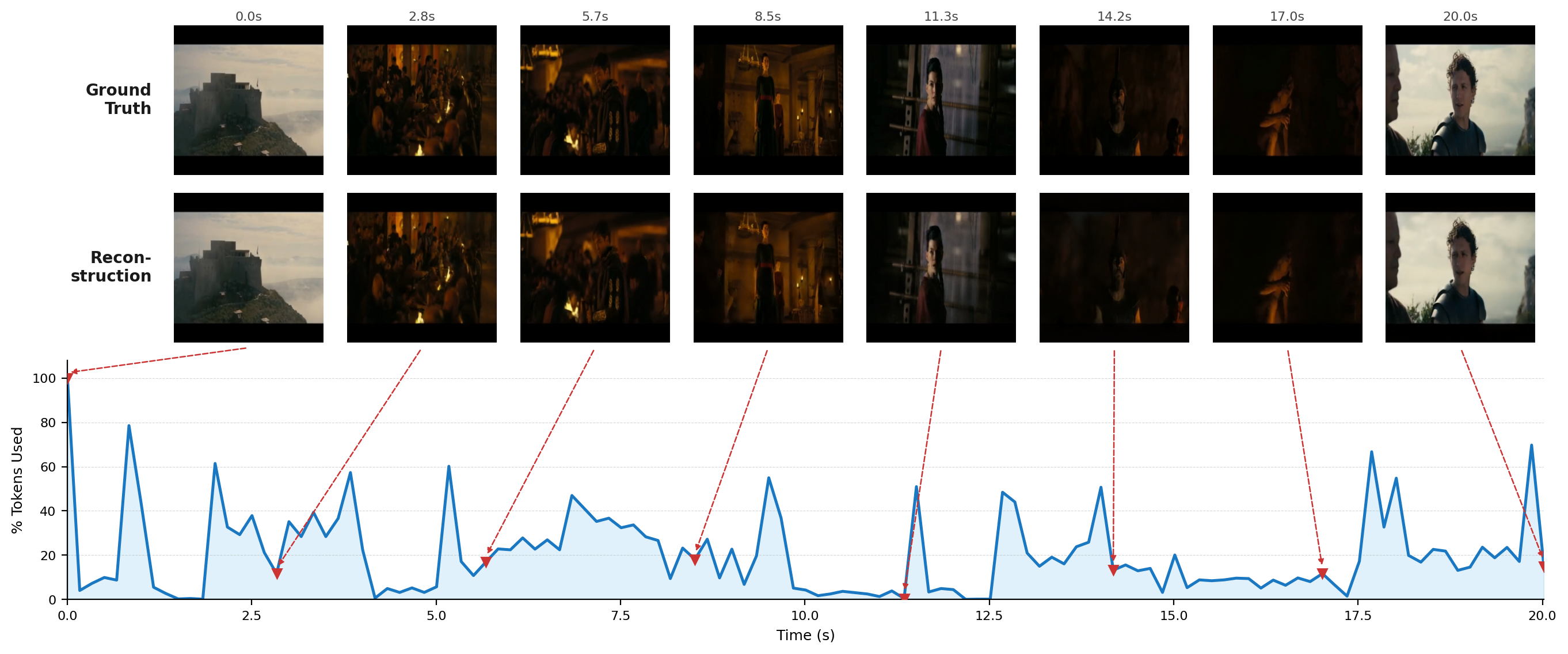}
    \caption{
        Per-frame token usage on representative Odyssey clips using our cached method ~\ref{sec:exp:cached-ref}. Top
        row shows the input video frames; bottom row plots the
        fraction of latent positions retained by our method as a
        function of temporal position at $\tau = 0.3$. Static
        intervals (visually stationary frames) retain few tokens;
        dynamic intervals (motion, scene change) retain more. The
        threshold $\tau$ is fixed across the entire clip, yet the
        retained-token count varies meaningfully over time,
        demonstrating that the criterion captures genuine
        content-driven variation rather than imposing a uniform rate.
    }
    \label{fig:token-usage}
\vspace{-1.2em}
\end{figure*}

There have been recent attempts at addressing this bottleneck via \emph{adaptive tokenisation}. ElasticTok \cite{yan2025elastictok} trains an auto-encoder with a mask sampled from a uniform distribution, applied on the \textit{tail} \cite{hu2024matryoshkaquerytransformerlarge} of the tokens, and supports four mechanisms for selecting per-input token counts at inference: Exhaustive search, Binned search over uniformly spaced lengths, Binary search assuming monotonic reconstruction loss, or a neural regression fine-tuned on paired (input, length) examples. AdapTok \cite{adaptok} trains a block level scorer and an integer linear programming allocator (\textbf{IPAL}) that jointly determines token counts per video. InfoTok ~\cite{infotok} grounds adaptive allocation in Shannon's source coding theorem, deriving an Evidence lower bound (\textbf{ELBO}) based router that allocates tokens by content complexity.

These methods work. However, each of them introduce marginal overheads. ElasticTok's inference-time mechanisms require several decoder evaluations per input; the neural regressor requires a trained scorer plus an inference time optimisation step. InfoTok requires an additional full-rate decoder pass to estimate per-block ELBO before allocating tokens, effectively doubling the inference cost. The added complexity is justified by the rate-quality gains, but a natural question then emerges, "is it necessary?"

To answer this, we shift focus to the continuous tokenisation regime. While most prior adaptive work \cite{infotok, adaptok} operates in discrete spaces, generative modelling is pivoting toward continuous latents to avoid quantisation bottlenecks: driven by Latent Video Diffusion Models \cite{blattmann2023alignlatentshighresolutionvideo, guo2024animatediffanimatepersonalizedtexttoimage} and large-scale generators like Sora \cite{liu2024sorareviewbackgroundtechnology}, tokenisers such as CV-VAE \cite{zhao2024cvvaecompatiblevideovae} and Cosmos \cite{agarwal2025cosmos} preserve the fine-grained gradients essential for high-fidelity synthesis \cite{deng2025autoregressivevideogenerationvector}. Crucially, this regime offers a largely unexplored advantage for adaptive tokenisation: unquantised latent vectors natively encode geometric distance.

The latent space of a frozen continuous video tokeniser already contains the signal needed to identify the redundant token positions: When consecutive temporal positions at the same spatial location produce nearly identical latent vectors, the second position adds little information. Detecting this requires no scoring network, and no extra decoder pass. It is simply a comparison between two latent vectors. The idea has a precedent in pixel-space: Run-length tokenisation (RLT) ~\cite{rlt}, originally proposed to accelerate video transformers, prune redundant patches by thresholding pixel-space L1 differences between consecutive frames. We transfer this principle to the latent space of a continuous video tokeniser.

Our pipeline has three components. A frozen continuous video tokeniser encoder produces latents from the input video. A masking criterion computes per-position temporal-L1 differences in the latent space and drops positions whose differences fall below a fixed threshold $\tau$. A Latent Inpainting Transformer (LIT) reconstructs the dropped positions via factorised spatial and temporal attention over the retained tokens. The frozen decoder then produces the final reconstruction. Crucially, the threshold $\tau$ is a single scalar whose magnitude is dictated primarily by the inherent variance of the chosen backbone's latent space, calibrated just once to establish a nominal operating point. 

The compression rate emerges from the complexity of the input rather than being specified
up front. Static scenes produce many sub-threshold differences and
compress aggressively, while dynamic scenes retain more tokens. On UCF-101 at
$\tau = 0.3$, per-video drop rates range from 5.15\% to
86.10\%, reflecting genuine content-driven variation.

\paragraph{Contributions}
\begin{itemize}
    \item We show that thresholding temporal-L1 differences in the latent
    space of a frozen tokeniser provides an adequate  signal for
    adaptive video tokenisation, eliminating the need for learned
    routers, scoring networks, or extra decoder passes at inference
    time.
    
    \item We introduce LIT, a factorised transformer that reconstructs
    dropped latent positions through interleaved spatial and temporal
    attention with 2D rotary positional embeddings.
    
    \item On TokenBench and DAVIS, we demonstrate that our threshold-driven approach achieves reconstruction fidelity competitive with the state-of-the-art continuous adaptive baseline, ElasticTok-CV \cite{yan2025elastictok}, while utilising fewer tokens, and drastically reducing compute overhead. For completeness, we also establish favourable performance against the recent discrete adaptive method, namely InfoTok \cite{infotok}, under identical retention constraints. Furthermore, the emergent per-video keep rate varies meaningfully with motion complexity, confirming genuine content-adaptive compression rather than a uniformly imposed rate.
\end{itemize}

\begin{figure*}[t]
    \centering
    \includegraphics[width=0.88\textwidth]{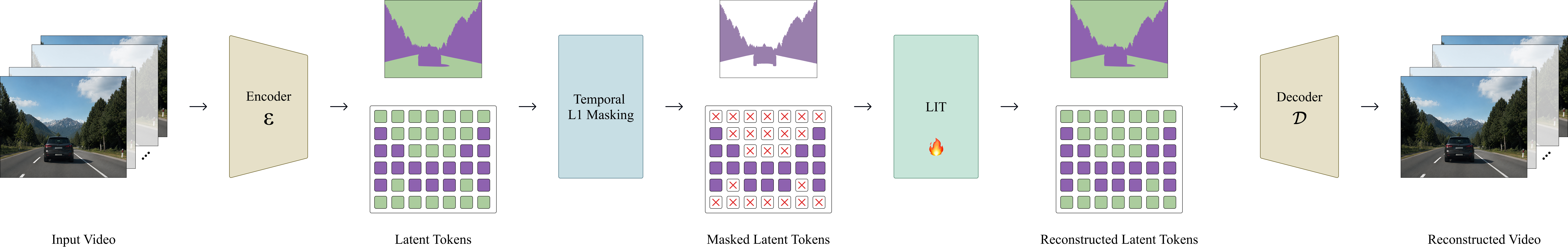}
    \caption{
        A frozen continuous video tokeniser encoder
        $\mathcal{E}$ maps an input clip $x$ to a latent
        $z \in \mathbb{R}^{C \times t \times h \times w}$. The
        temporal-L1 masking module computes per-position differences
        between consecutive temporal positions under the
        \emph{last-kept} reference scheme and produces a binary keep
        mask $m$ via thresholding at $\tau$. Positions below the
        threshold are dropped (zeroed). The Latent Inpainting
        Transformer (LIT) recovers the dropped positions via
        interleaved spatial and temporal attention with 2D rotary
        positional embeddings, producing $\hat{z}$. The frozen
        decoder $\mathcal{D}$ then reconstructs the video $\hat{x}$
        from $\hat{z}$. Only LIT is trained; token allocation requires no
        learned parameters and no extra decoder pass at inference.
    }
    \label{fig:architecture}
\vspace{-1.2em}
\end{figure*}



\section{Related Work}

\subsection{Discrete and Continuous Video Tokenisation}

The modern lineage of learned visual tokenisation begins with VQ-VAE~\cite{van2017neural}, which maps image patches to entries in a jointly learned finite codebook, and VQ-GAN~\cite{esser2021taming}, which added adversarial training and showed discrete tokens sufficient for high-fidelity autoregressive synthesis. For video, MAGVIT~\cite{magvit} extended the 3D convolutional VQ framework to class-conditional generation, while MAGVIT~v2~\cite{yu2024language} showed that tokeniser quality, rather than generative model capacity, is the decisive factor in downstream fidelity.

Continuous tokenisers forgo codebooks entirely, mapping video to a compact latent grid without quantisation error. OmniTokenizer~\cite{omnitokenizer} proposed a joint image-video tokeniser with shared parameters, and LARP~\cite{Wang2024LARPTV} introduced holistic latent tokens encoding global video information through a learned autoregressive prior. The Cosmos Tokenizer~\cite{agarwal2025cosmos}, which we adopt as our backbone, achieves $4\times 8\times 8$ spatio-temporal compression over a 16-channel continuous latent space and represents the current state of the art. Across both regimes, however, these works apply a fixed compression rate to every input irrespective of content, a rigid bottleneck our work explicitly addresses.

\subsection{Adaptive Tokenisation}
The observation that token budgets should reflect informational complexity has motivated a growing body of work on adaptive tokenisation. ElasticTok~\cite{yan2025elastictok} introduced the first
adaptive video tokeniser, applying a random tail-drop masking scheme during training that imposes a positional ordering: early tokens carry global structure, late tokens carry fine detail that can be sacrificed under compression. Inference requires a binary search over token lengths, incurring approximately $\log_2 N$ forward passes. AdapTok~\cite{adaptok} moved to a causal 1D latent space, decoupling token ordering from spatial structure, and introduced an Integer Linear Programming allocation strategy that jointly optimises token budgets across a batch. InfoTok~\cite{infotok} provided the first theoretical
grounding for adaptive tokenisation: via Shannon's Source Coding Theorem, it proved that position-based tail-drop methods are provably biased, and proposed an Evidence Lower Bound (ELBO) router that assigns token budgets proportional to per-video reconstruction difficulty, requiring only two forward passes at inference. KARL~\cite{Duggal2025SinglepassAI} addressed inference efficiency for images by framing the token count
as an approximation to Kolmogorov Complexity and predicting it in a single pass through a learned halting head.

Our method occupies a distinct point in this design space. We operate in the continuous-latent regime, complementing the discrete-codebook line of work \cite{adaptok, infotok}. And rather than learning a budget routing network or imposing a positional ordering on tokens, we derive a fully parameter-free mask from the temporal structure of the latent representation itself, requiring no training of the masking criterion and no additional network evaluations to determine token budgets.

\subsection{Token Redundancy and Temporal Pruning}

Closest to our approach is Run-Length Tokenisation (RLT)~\cite{rlt}, which exploits consecutive temporal redundancy in pixel space to prune vision-transformer inputs: tokens whose pixel values have not changed appreciably since the most recently retained frame are dropped, on the basis that they can be recovered from their temporal neighbours. Our work transfers this insight to the continuous latent space of a frozen video tokeniser. Operating in latent space rather than pixel space offers two advantages: the latent representation is already spatially compressed, and the frozen backbone's inductive biases ensure that temporally correlated
pixel-space content produces temporally correlated latent vectors. We further complement the parameter-free masking criterion with an inpainting network that explicitly reconstructs dropped positions from their retained neighbours, recovering information that pixel-space pruning methods discard entirely.


\section{Method}
We present a video tokenisation framework that achieves content-adaptive compression without learned budget routing. The framework comprises three components:
a frozen continuous video tokeniser backbone that provides the latent representation, a parameter-free temporal-redundancy masking criterion that
determines which latent positions to drop, and a factorised transformer network that reconstructs the full latent from the masked subset.
We describe each component below and conclude with the training objective. The exact mechanisms of our method's forward pass and inference can be found in Algorithm~\ref{alg:lit_pipeline}.

\begin{algorithm}[!htbp]
\caption{Latent Inpainting Transformer (LIT) Pipeline}
\label{alg:lit_pipeline}
\begin{algorithmic}[1]
    \Require Frozen encoder $\mathcal{E}$; frozen decoder $\mathcal{D}$; threshold $\tau$; LIT parameters $\theta$
    
    \Statex \textbf{\textit{Stage 1: Training}}
        \Require Video dataset $\mathcal{X}$; loss weights $\lambda_{\text{recon}}, \lambda_{\text{latent}}$
        \For{each training step}
            \State Sample minibatch $\{x_i\}_{i=1}^{B} \subset \mathcal{X}$
            \State $z_i \gets \mathcal{E}(x_i)$ \Comment{Encode with frozen encoder}
            \State $m_i \gets \text{TemporalL1Mask}(z_i, \tau)$ \Comment{Algorithm~\ref{alg:temporal-l1}}
            \State $\tilde{z}_i \gets z_i \odot m_i$ \Comment{Drop redundant positions}
            \State $\hat{z}_i \gets \text{LIT}_\theta(\tilde{z}_i)$ \Comment{Reconstruct latent}
            \State $\hat{x}_i \gets \mathcal{D}(\hat{z}_i)$ \Comment{Decode with frozen decoder}
            \State $\mathcal{L} \gets \lambda_{\text{recon}} \lVert x_i - \hat{x}_i \rVert^2_2 + \lambda_{\text{latent}} \lVert z_i - \hat{z}_i \rVert^2_2$
            \State Update $\theta$ via gradient descent on $\mathcal{L}$
        \EndFor
        
    \Statex \hrulefill 
    \Statex \textbf{\textit{Stage 2: Inference}}
    \Require Input video $x$, threshold $\tau$
    \State $z \gets \mathcal{E}(x)$ \Comment{Single encoder pass}
    \State $m \gets \text{TemporalL1Mask}(z, \tau)$ \Comment{Parameter-free mask}
    \State $\tilde{z} \gets z \odot m$ \Comment{Keep only $\sum m$ tokens}
    \State $\hat{z} \gets \text{LIT}_\theta(\tilde{z})$
    \State $\hat{x} \gets \mathcal{D}\!\left(\hat{z}\right)$
    \State \Return $\hat{x}$
\end{algorithmic}
\end{algorithm}

\subsection{Continuous Video Tokeniser Backbone}
\label{sec:method:backbone}
We build on Cosmos-Tokenize1-CV 4$\times$8$\times$8 \cite{agarwal2025cosmos}, a continuous tokeniser with no learned quantisation. However, our method can be used for any continuous video tokeniser. Cosmos' backbone consists of an encoder-decoder pair $(\mathcal{E}, \mathcal{D})$ that we treat as frozen throughout. The encoder $\mathcal{E}: \mathbb{R}^{3\times T \times H \times W}$ maps a video clip of $T$ frames, having a spatial resolution of $H \times W$ to a continuous latent tensor with $C = 16$ channels and spatio temporal dimensions $(t, h, w) = (\lceil T/ 4 \rceil, H/8, W/8)$ reflecting the backbone's $4 \times 8 \times 8$ compression. The decoder $\mathcal{D}$ acts as the inverse, inverting the encoding.

We adopt $T=33$ and $H = W = 256$ as our standard clip configuration, following the convention of recent adaptive tokeniser \cite{infotok, agarwal2025cosmos}. This returns a latent grid of shape $(16, 9, 32, 32)$ where $N = thw = 9216$ represents the tokens per clip.

Throughout, $\mathcal{E}$ and $\mathcal{D}$ remain frozen i.e., no gradients are propagated through them. Our trainable components operate entirely on the latent representation $z = \mathcal{E}(x) \in \mathbb{R}^{C \times t \times h \times w}$.

\subsection{Temporal Latent Masking}
\label{sec:method:masking}

Inspired by Run-Length tokenisation \emph{(RLT)} \cite{rlt}, which exploits redundancy in pixel-space for vision transformer pruning, we adopt a parameter free criterion for selecting which latent positions to drop. The motivating observation, shared with RLT, is the consecutive temporal positions at the same spatial location differ only slightly when the scene is locally stationary. Dropping such tokens impose minimal information loss, since the receiver can recover them approximately from their retained neighbours. Our
contribution lies in transferring this insight to the latent space of a
frozen continuous video tokeniser and demonstrating that a learned
inpainting network can effectively reconstruct the dropped positions

\begin{figure*}[t]
    \centering
    \includegraphics[width=0.90\linewidth]{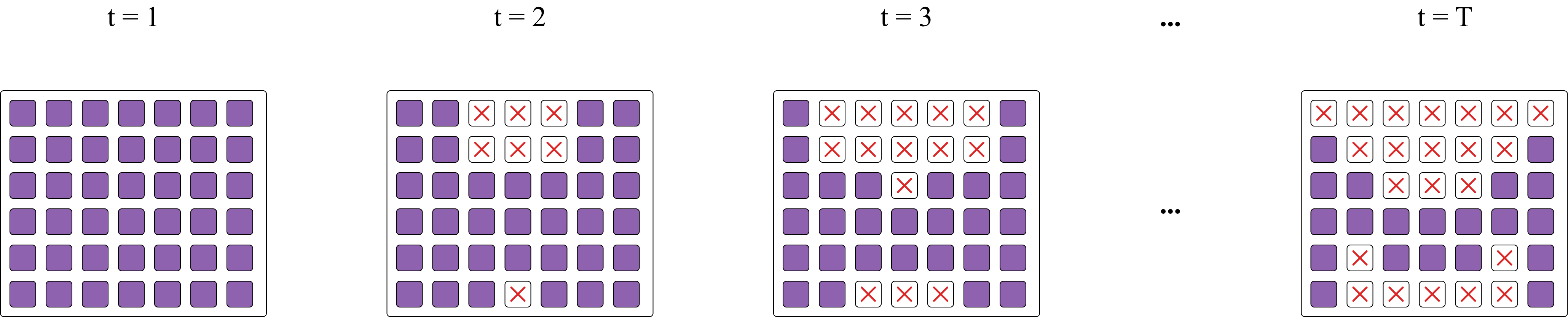}
    \caption{
        A Toy example illustrating the temporal-L1 masking under the \emph{last-kept} reference scheme. For each spatial position $(y, x)$, the algorithm
        maintains a reference latent corresponding to the most recently retained temporal position. At each subsequent
        temporal position $t'$, the per-channel L1 difference $\Delta$ between the current latent and the reference is
        computed and compared against threshold $\tau$. Positions with $\Delta \geq \tau$ are retained (and become the new
        reference); positions with $\Delta < \tau$ are dropped. The first temporal position is always retained as the initial
        reference. This produces a content-adaptive mask whose density follows the input's temporal redundancy.
    }
    \label{fig:masking}
\vspace{-1.2em}
\end{figure*}

\subsection{The Masking Criterion}
\label{sec:method:masking-criterion}
Let $z[:, t', y, x] \in \mathbb{R}^{C}$ denote the latent vector at temporal position $t'$ and spatial position $(y, x)$. For each spatial position, we
process temporal positions sequentially and maintain a \emph{reference}: the most recently retained latent at this spatial position. Formally, define
$\rho(y, x, t') = \max\{t'' < t' : m_{t'',y,x} = 1\}$ as the index of the most recently retained temporal position at $(y, x)$ prior to $t'$.

The temporal L1 difference at position $(t', y, x)$ for $t' > 0$ is 
\begin{equation}
    \Delta(t', y, x) = \frac{1}{C} \sum_{c=1}^C \bigl| z[c, t', y, x] - z[c, \rho(y, x, t'), y, x] \bigr|
    \label{eq:l1-diff}
\end{equation}

An empirical analysis of the Cosmos latent space revealed minimal cross-channel variance. Consequently, taking the channel-wise mean L1 distance provides a highly stable heuristic for capturing temporal redundancy, circumventing the need for more complex normalisation or channel-wise maximums.

The first temporal position at each spatial location is always retained, providing an unmasked reference. For subsequent positions, we retain the position if its latent has changed by more than $\tau$ relative to the most recently retained version at the same location.

\subsection{Algorithm and Properties}
\label{sec:method:masking-algorithm}

The mask is computed by a single causal pass through the temporal axis. Algorithm~\ref{alg:temporal-l1} specifies the procedure. The computation requires only the encoded latent and admits batched implementation on accelerators.

\begin{algorithm}[t]
\caption{Temporal-L1 latent masking.}
\label{alg:temporal-l1}
\begin{algorithmic}[1]
    \Require Latent $z \in \mathbb{R}^{C \times t \times h \times w}$, threshold $\tau$
    \Ensure Keep mask $m \in \{0, 1\}^{t \times h \times w}$
    \State $m[0, :, :] \gets 1$ \Comment{always keep first temporal position}
    \State $\text{ref} \gets z[:, 0, :, :]$ \Comment{reference latent for each $(y, x)$}
    \For{$t' = 1$ to $t - 1$}
        \State $\Delta \gets \frac{1}{C} \sum_{c} \lvert z[c, t', :, :] - \text{ref}[c, :, :] \rvert$ \Comment{shape $(h, w)$}
        \State $m[t', :, :] \gets \mathds{1}[\Delta \geq \tau]$
        \State $\text{ref} \gets m[t', :, :] \cdot z[:, t', :, :] + (1 - m[t', :, :]) \cdot \text{ref}$
    \EndFor
    \State \Return $m$
\end{algorithmic}
\end{algorithm}

With our method, there are three properties which warrant emphasis.
\paragraph{Content-adaptive compression}
The fraction of tokens retained depends on the temporal structure of the video: static camera footage compresses aggressively, whereas footage with substantial motion produces differences that mostly exceed $\tau$. The threshold itself is fixed, in contrast to rate-based methods that fix the compression rate $r$ as a hyperparameter and then decide which tokens to drop. Qualitative visualisations of these emergent masks across diverse samples are provided in \hyperref[sec:appendix-b]{Appendix~B} in the Supplementary material.

\paragraph{Parameter free}
We use \emph{parameter-free} in a precise sense: no \emph{learned} component participates in the masking decision and no search or router is evaluated to obtain a budget. It does not mean the criterion is free of hyperparameters; it has exactly one, the threshold $\tau$, a single interpretable scalar calibrated once per backbone and then held fixed (Section~\ref{sec:method:tau-calibration}). This contrasts with prior work which trains a router network \cite{adaptok} or maintains running dataset statistics to guide allocation \cite{infotok}.

\paragraph{Single pass}
Computing the mask requires only the latent $z$, already produced by the encoder; no additional decoder pass is needed at inference. Concretely, the forward evaluations spent on token allocation number zero, against one for InfoTok and $\approx \log_2 N$ for ElasticTok's binary search \cite{yan2025elastictok}. This is the sense in which we claim an efficiency advantage, and it is what the wall-clock measurements of Table~\ref{tab:nfe} reflect.

\paragraph{The compressed representation}
The dense zeroed grid $\tilde{z} = z \odot m$ is an implementational convenience, not the transmitted object: the representation is the set of \emph{retained} latents plus the binary mask $m$, realised as a gather over $z$, so its size scales with $\sum m$ rather than $thw$. The dense grid is reconstituted transiently at the decoder endpoint.


\subsection{Choosing the Threshold $\tau$}
\label{sec:method:tau-calibration}

The threshold $\tau$ controls the amount of temporal redundancy removed in latent space. We perform $\tau$ calibration via an empirical sweep, inspecting both the generated masks and the resulting reconstruction quality across candidate thresholds.

For the Cosmos backbone, we found $\tau = 0.3$ to provide the best trade-off between compression and fidelity. Lower values retain a large number of temporally redundant background tokens, while higher values begin to suppress tokens corresponding to genuine motion.

$\tau$, the criterion's only hyperparameter, traces a rate--quality curve along which quality degrades smoothly and monotonically, with no collapse threshold and no narrow basin to locate by search (\hyperref[sec:appendix-a]{Appendix~A} in the Supplementary material). The operating point is backbone-dependent (\hyperref[sec:appendix-e]{Appendix~E} in the Supplementary material), but a single $\tau = 0.3$ serves every Cosmos result here, on TokenBench and DAVIS alike.

On UCF-101, $\tau = 0.3$ yields an average token keep rate of $47.78\%$ under the last-kept reference scheme (Equation~\ref{eq:l1-diff}). The per-video drop rate varies from $5.15\%$ on the least redundant clips to $86.10\%$ on the most redundant clips, demonstrating that the resulting token budget adapts naturally to video content. Further details of the calibration procedure are provided in \hyperref[sec:appendix-a]{Appendix~A} in the Supplementary material.





\subsection{Latent Inpainting Transformer (LIT)}
\label{sec:method:lit}

Given the masked latent $\tilde{z} = z \odot m$, the network's task is to approximate $\hat{z}$ of the full-latent $z$. We frame this as an inpainting problem in the latent space where kept tokens provide information from which the dropped tokens must be inferred.

A direct approach is full 3-D self-attention over all $thw$ tokens, which costs $O((thw)^2 D)$ per layer, where $D$ is the hidden dimension, and scales prohibitively in both compute and memory. Factorised attention instead decomposes this into separate interactions: spatial attention restricted to tokens sharing a temporal slice, giving $t$ independent computations of size $hw$, and temporal attention restricted to tokens sharing a spatial column, giving $hw$ independent computations of size $t$. The combined cost is $O\bigl(t \cdot (hw)^2 D + hw \cdot t^2 D\bigr)$, which for our dimensions $(t, h, w) = (9, 32, 32)$ amounts to a reduction of approximately $9\times$ ($9.5 \times 10^6$ against $8.5 \times 10^7$). Factorisation also supplies a useful inductive bias: spatial textures and temporal motion patterns have distinct statistics, and each branch can specialise to one of them.

\subsection{Architecture}
\label{sec:method:lit-architecture}

The Latent Inpainting Transformer (LIT) is an architecture with alternating spatial and temporal
attention blocks. Let $h^{(\ell)} \in \mathbb{R}^{thw \times D}$ denote
the hidden representation at layer $\ell$, where $D$ is the hidden
dimension. The initial representation $h^{(0)}$ is obtained by linearly
projecting the masked latent: $h^{(0)} = \text{InProj}(\tilde{z}) + \text{PE}$,
where $\text{PE}$ is the positional encoding described in
Section~\ref{sec:method:rope}.

A \emph{spatial block} applies the following operations:
\begin{equation}
    \begin{aligned}
        h^{(\ell)}_{\text{attn}} &= h^{(\ell-1)} + \text{SpatialMHA}\bigl(\text{LN}(h^{(\ell-1)})\bigr), \\
        h^{(\ell)} &= h^{(\ell)}_{\text{attn}} + \text{FFN}\bigl(\text{LN}(h^{(\ell)}_{\text{attn}})\bigr),
    \end{aligned}
    \label{eq:spatial-block}
\end{equation}
where SpatialMHA performs multi-head self-attention restricted to
within-slice token pairs (i.e., tokens sharing the same $t'$).

A \emph{temporal block} applies the analogous operations along the
temporal axis:
\begin{equation}
    \begin{aligned}
        h^{(\ell)}_{\text{attn}} &= h^{(\ell-1)} + \text{TemporalMHA}\bigl(\text{LN}(h^{(\ell-1)})\bigr), \\
        h^{(\ell)} &= h^{(\ell)}_{\text{attn}} + \text{FFN}\bigl(\text{LN}(h^{(\ell)}_{\text{attn}})\bigr),
    \end{aligned}
    \label{eq:temporal-block}
\end{equation}
where TemporalMHA performs attention restricted to within-column token
pairs (sharing the same $(y, x)$). Both blocks use pre-LayerNorm and the
standard transformer FFN with $4 \times D$ expansion and GELU activation.

The network stacks $L_{\text{main}} = 6$ pairs of (spatial, temporal)
blocks, followed by $L_{\text{ref}} = 3$ refinement pairs that further
process the representation. After the final block, a linear projection
maps the hidden representation back to the latent dimension:
\begin{equation}
    \hat{z} = \text{OutProj}(h^{(L)}).
    \label{eq:out-proj}
\end{equation}

We instantiate LIT with hidden dimension $D=192$, $H = 8$ attention heads, $L_{main} = 6$ main blocks and $L_{ref} = 3$ refinement blocks, totalling approximately 2.7 million trainable parameters.

\subsubsection{Positional Encoding via RoPE}
\label{sec:method:rope}

We use 2D Rotary Position Embeddings (RoPE)~\cite{su2023roformerenhancedtransformerrotary} to encode
spatiotemporal positions. For spatial attention layers, RoPE is applied
along the $(y, x)$ axes, encoding the spatial position of each token
within its temporal slice. For temporal attention layers, RoPE is applied
along the $t'$ axis, encoding the temporal position of each token within
its spatial column. The use of RoPE removes the need for learnable
positional embedding tables and provides relative-position structure that
generalises across token-count variations.

This choice differs from RLT~\cite{rlt}, which augments tokens with a
learnable length bias representing the number of consecutive frames a
surviving token represents. Their bias is well-suited to downstream
classification tasks, where the run-length informs how strongly to weight
a token in the final prediction. For our task the run length information isn't required, the LIT's objective is purely local (fill-in missing tokens from neighbours).

\subsubsection{Reconstruction}
\label{sec:method:lit-output}

After the LIT produces $\hat{z}$, the final reconstruction is
obtained by passing $\hat{z}$ through the frozen Cosmos decoder:
\begin{equation}
    \hat{x} = \mathcal{D}(\hat{z}).
    \label{eq:reconstruction}
\end{equation}
The LIT network is the only learned component in the inference pipeline.
Its task is to approximately invert the masking operation, producing a
latent that the frozen decoder can decode into a high-quality
reconstruction.

\subsection{Training Objective}
\label{sec:method:training}

We train the LIT by minimising a combination of pixel-space and
latent-space reconstruction losses:
\begin{equation}
    \mathcal{L} = \lambda_{\text{recon}} \, \mathcal{L}_{\text{recon}}(x, \hat{x}) + \lambda_{\text{latent}} \, \mathcal{L}_{\text{latent}}(z, \hat{z}),
    \label{eq:loss}
\end{equation}
where
\begin{equation}
    \mathcal{L}_{\text{recon}}(x, \hat{x}) = \lVert x - \hat{x} \rVert^2_2, \qquad
    \mathcal{L}_{\text{latent}}(z, \hat{z}) = \lVert z - \hat{z} \rVert^2_2.
    \label{eq:loss-terms}
\end{equation}
The latent-space term provides direct supervision on the LIT's
primary output, while the pixel-space term ensures that reconstruction
quality is measured against the original video content. We set
$\lambda_{\text{recon}} = 1.0$ and $\lambda_{\text{latent}} = 1.0$ in our
main experiments.

\section{Experiments}
\label{sec:experiments}

We evaluate on standard video reconstruction benchmarks against fixed-rate and adaptive tokenisation baselines. Our experiments address three questions: (1) does temporal-L1 thresholding in the latent space produce a useful budget routing signal? (2) how does the method compare to existing adaptive tokenisers? and (3) what is the inference-cost advantage over methods requiring extra decoder evaluations?

\subsection{Setup}
\label{sec:exp:setup}

We train on UCF-101~\cite{soomro2012ucf101dataset101human} and Kinetics-400~\cite{kay2017kinetics}, and
evaluate on TokenBench~\cite{nvidia2024cosmos}, and DAVIS \cite{caelles20192019davischallengevos}. We use Cosmos-Tokenize1-CV4$\times$8$\times$8~\cite{agarwal2025cosmos}
as the frozen backbone for all experiments. Videos are sampled as
33-frame clips at $256 \times 256$ resolution with frame interval~3,
yielding latent grids of shape $(16, 9, 32, 32)$ with $N = 9{,}216$
tokens per clip. We train LIT with AdamW \cite{loshchilov2019decoupledweightdecayregularization} at learning rate
$5 \times 10^{-4}$, weight decay $10^{-4}$, cosine annealing to
$10^{-5}$, and effective batch size~16. We train for $10{,}000$ steps
and select checkpoints by validation PSNR.
We report PSNR, SSIM~\cite{ssim}, LPIPS-VGG~\cite{zhang2018unreasonableeffectivenessdeepfeatures} and Frechet Video Distance (FVD)~\cite{unterthiner2019accurategenerativemodelsvideo} as reconstruction-quality metrics, together with the \emph{keep rate}, the fraction of latent positions retained, as the compression axis: absolute bit-cost differs between continuous and discrete tokenisers and is not comparable across regimes. For fair comparison across tokenisers we resize all videos to $H = W = 256$, with 33-frame clips at frame interval~1.

\subsection{Inference Cost}
\label{sec:exp:cost}

A key practical advantage of our method is its single-pass inference. Table ~\ref{tab:nfe} illustrates the number of forward evaluations required by each adaptive tokenisation method, and the corresponding wall-clock measurements on a single NVIDIA A10G GPU at fixed input resolution.

\begin{table}[!hbtp]
\centering
\small
\setlength{\belowcaptionskip}{8pt}
\caption{
    Per-clip inference cost. NFE counts the additional forward evaluations for computing the token budget. ElasticTok and our
    method operate in the continuous regime; InfoTok is included
    despite operating in the discrete regime. The Speed up is computed relative to ElasticTok-CV.
}
\label{tab:nfe}
\setlength{\tabcolsep}{2pt}
\begin{tabular}{@{}lccc@{}}
\toprule
Method & NFE $\downarrow$ & Per-clip (secs) $\downarrow$ & Speed Up  $\uparrow$\\
\midrule
ElasticTok-CV (binary search) & $\log_2 N$ & 35.97 & 1$\times$ \\
InfoTok (discrete) & 1 & 2.84 & 13$\times$ \\
\midrule
\textbf{Ours} & 0 & 1.15 & 31$\times$ \\
\bottomrule
\end{tabular}
\end{table}

\subsection{Evaluation Protocol and Baseline Constraints}
Comparing adaptive tokenisers with distinct rate-control mechanisms is empirically challenging. Our Temporal-L1 method yields compression as an emergent property of motion complexity, whereas ElasticTok~\cite{yan2025elastictok} uses iterative search, and InfoTok~\cite{infotok} targets discrete budgets via ELBO. To ensure rigorous comparison:

\noindent\textbf{Iso-Quality (vs.\ ElasticTok-CV):} Positional tail-dropping degrades spatial details at low budgets. We thus constrain ElasticTok by a strict error threshold (MSE = 0.003). A detailed overview of the results can be found in Table~\ref{tab:unified_main_results}.

\noindent\textbf{Iso-Rate (vs.\ InfoTok):} For discrete baselines sharing the Cosmos grid, we constrain their budget parameter ($\beta$) to approximate our emergent keep rate. While discrete indices naturally compress further in absolute bit-cost, this isolates and evaluates pure allocation efficacy.

\subsection{Results}
As shown in Table~\ref{tab:unified_main_results}, we evaluate the reconstruction performance of our method across both TokenBench and DAVIS datasets. Our framework demonstrates exceptional compression efficiency and generalization. On TokenBench, our Cosmos-based variant retains only 32\% of spatial-temporal tokens yet vastly outperforms InfoTok at the same budget (+2.81~dB PSNR, -107.55 FVD). This efficiency extends to the DAVIS dataset, where our model dynamically scales its emergent keep rate to 62\%, outperforming InfoTok by +2.47~dB PSNR with superior temporal consistency (312.66 vs. 406.33 FVD). While ElasticTok-CV achieves slightly higher absolute metrics on DAVIS, it demands a substantially higher token budget (82\% keep rate), positioning our framework as a more optimal solution for aggressive data reduction without catastrophic fidelity loss.

The OmniTokenizer variant shows the criterion transfers to a second continuous backbone at a $29\%$ keep rate, though with narrower margins. Since the criterion is identical, the difference originates in the latent spaces: \hyperref[sec:appendix-e]{Appendix~E} in the Supplementary material analyses this via their \emph{scale} and \emph{selectivity} of the score distribution.

\begin{table*}[h] 
\centering
\caption{Reconstruction quality across diverse datasets. To ensure fair evaluation across distinct rate-control mechanisms, InfoTok is constrained to approximate our emergent keep rate, while ElasticTok-CV is constrained by a strict reconstruction error threshold (MSE = 0.003). Keep rate indicates the fraction of latent spatial-temporal positions retained.}
\vspace{0.4em}
\label{tab:unified_main_results}
\resizebox{\textwidth}{!}{
\begin{tabular}{l ccccc ccccc}
\toprule
& \multicolumn{5}{c}{\textbf{TokenBench (256x256)}} & \multicolumn{5}{c}{\textbf{DAVIS (256x256)}} \\
\cmidrule(lr){2-6} \cmidrule(lr){7-11}
Method & $\text{Keep Rate}\downarrow$ & $\text{PSNR}\uparrow$ & $\text{SSIM}\uparrow$ & $\text{LPIPS}\downarrow$ & $\text{FVD}\downarrow$ & $\text{Keep Rate}\downarrow$ & $\text{PSNR}\uparrow$ & $\text{SSIM}\uparrow$ & $\text{LPIPS}\downarrow$ & $\text{FVD}\downarrow$ \\
\midrule
Cosmos-CV4x8x8 (Fixed) & 100\% & 35.38 & 0.948 & 0.078 & 8.77 & 100\% & 31.57 & 0.918 & 0.109 & 61.33 \\
OmniTokenizer-VAE (Fixed) & 100\% & 27.05 & 0.892 & 0.128 & 57.83 & 100\% & 24.49 & 0.837 & 0.194 & 244.81 \\
\midrule
ElasticTok-CV (MSE=0.003) & 40\%* & 30.87 & 0.900 & 0.154 & 68.75 & 82\%* & 29.02 & 0.878 & 0.172 & 200.97 \\
InfoTok (Discrete) & 32\%* & 27.50 & 0.817 & 0.234 & 189.16 & 62\%* & 25.80 & 0.787 & 0.244 & 406.33 \\
\midrule
\textbf{Ours (Cosmos, }$\bm{\tau=0.3}$\textbf{)} & \textbf{32\%} &  \textbf{30.31} &  \textbf{0.894} &  \textbf{0.141} & \textbf{81.61} & \textbf{62\%} &  \textbf{28.27} & \textbf{0.856} & \textbf{0.187} & \textbf{312.66} \\
\textbf{Ours (Omni, }$\bm{\tau=1.2}$\textbf{)} & \textbf{29\%} & \textbf{23.88} & \textbf{0.790} & \textbf{0.245} & \textbf{168.82} & \textbf{52\%} & \textbf{22.50} & \textbf{0.736} & \textbf{0.293} & \textbf{468.15} \\
\bottomrule
\multicolumn{11}{p{\textwidth}}{\footnotesize *Baseline methods were strictly constrained by either an MSE threshold (ElasticTok) or target budget parameter (InfoTok)}
\end{tabular}
}
\end{table*}

\begin{figure*}[t]
    \centering
        \includegraphics[width=1\linewidth]{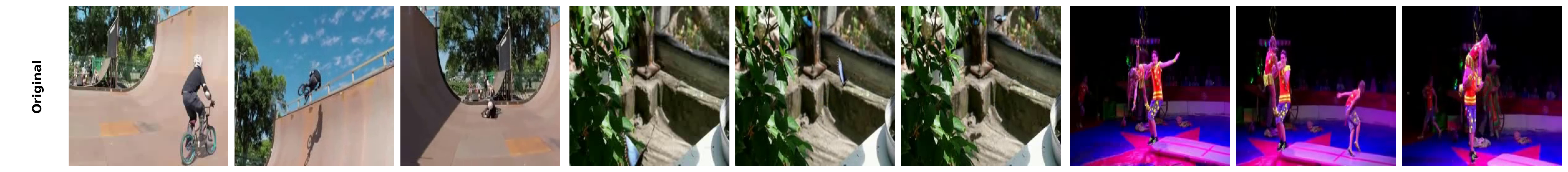}
        \vspace{0.5pt}
        \includegraphics[width=1\linewidth]{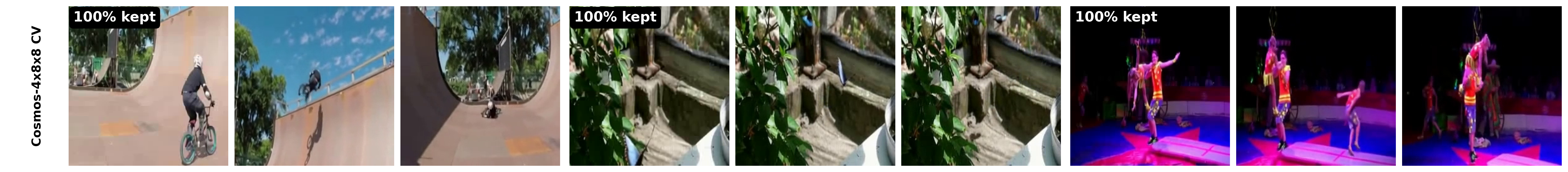}
        \vspace{0.5pt}
        \includegraphics[width=1\linewidth]{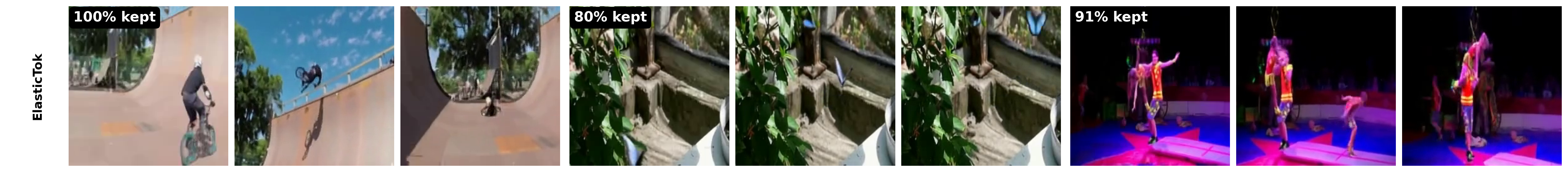}
        \vspace{0.5pt}
        \includegraphics[width=1\linewidth]{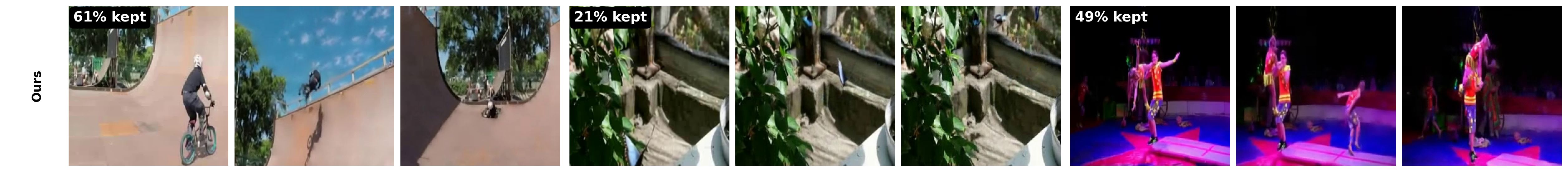}
    \caption{Reconstructions examples of video with different movements using different tokenisers. Our method can achieve similar reconstruction fidelity with higher compression compared to ElasticTok CV, and trivially, Cosmos-CV}
    \label{fig:placeholder}
\vspace{-1.2em}
\end{figure*}






\subsection{Ablations}
\label{sec:exp:ablations}

Our pipeline makes three independent choices: the criterion deciding retention, the reference it is measured against, and the fill applied to dropped positions. Table~\ref{tab:ablations} varies each in turn on DAVIS with the Cosmos backbone at a fixed $62\%$ keep rate.

\begin{table}[!hbtp]
\centering
\footnotesize
\caption{
    Ablations on DAVIS with the Cosmos backbone, all at a matched $62\%$ keep rate. Each block varies one component and holds the other two fixed at the setting named in its subheading. Rows marked \emph{ours} correspond to the configuration reported in Table~\ref{tab:unified_main_results}.
}
\vspace{0.4em}
\label{tab:ablations}
\setlength{\tabcolsep}{5pt}
\begin{tabular}{@{}lccccc@{}}
\toprule
Variant & Keep Rate & PSNR$\uparrow$ & SSIM$\uparrow$ & LPIPS$\downarrow$ & FVD$\downarrow$ \\
\midrule
\multicolumn{6}{@{}l}{\emph{(a) Criterion} \; --- \; reference: last-kept, \; fill: copy} \\
Pixel-L1 & 62\% & 25.59 & 0.775 & 0.261 & 629.25 \\
Latent-L2 & 62\% & 27.42 & 0.827 & 0.200 & \textbf{477.90} \\
Latent-cosine & 62\% & 26.14 & 0.814 & 0.209 & 509.28 \\
\textbf{Latent-L1 (ours)} & 62\% & \textbf{27.54} & \textbf{0.829} & \textbf{0.198} & 483.00 \\
\midrule
\multicolumn{6}{@{}l}{\emph{(b) Reference} \; --- \; criterion: latent-L1, \; fill: LIT} \\
Previous-frame & 62\% & 27.62 & 0.846 & 0.194 & 335.79 \\
\textbf{Last-kept (ours)} & 62\% & \textbf{28.27} & \textbf{0.856} & \textbf{0.187} & \textbf{312.66} \\
\midrule
\multicolumn{6}{@{}l}{\emph{(c) Fill} \; --- \; criterion: latent-L1, \; reference: last-kept} \\
Copy last-kept & 62\% & 27.54 & 0.829 & 0.198 & 483.00 \\
Linear interpolation & 62\% & 26.30 & 0.814 & 0.215 & 579.86 \\
\textbf{LIT (ours)} & 62\% & \textbf{28.27} & \textbf{0.856} & \textbf{0.187} & \textbf{312.66} \\
\bottomrule
\end{tabular}
\end{table}

Pixel-L1 is the weakest criterion ($-1.95$~dB PSNR, $+146$ FVD): even when its differences are pooled onto the tokeniser's exact spatial and temporal grid, change in RGB is a poor proxy for change in the latents being compressed. Within latent space the choice matters far less. The last-kept reference gains $+0.65$~dB over a previous-frame one, and the largest effect is the fill: replacing LIT with copying or interpolation costs $0.73$ and $1.97$~dB PSNR. Fidelity therefore comes from the inpainting stage, not the masking heuristic alone. \hyperref[sec:appendix-f]{Appendix~F} in the Supplementary material gives the full analysis.

\subsection{Streaming Cached Reference Across Clips}
\label{sec:exp:cached-ref}

The base masking criterion (Algorithm~\ref{alg:temporal-l1}) maintains its \emph{last-kept} reference only within a single 33-frame clip. When a longer video is processed as a sequence of consecutive chunks, this reference is reinitialised at every clip boundary. Consequently, the first latent frame of each new chunk must be retained, regardless of its similarity to the preceding chunk. This introduces periodic token-retention spikes that fail to exploit temporal continuity across chunk boundaries; \hyperref[sec:appendix-c]{Appendix~C} in the Supplementary material plots them against the streaming variant on a long video.

To address this limitation, we introduce a streaming variant that maintains two distinct caches across clips:

\begin{itemize}[noitemsep, topsep=0pt, partopsep=0pt, parsep=0pt]

\item \textbf{Selection Cache ($\mathbf{C}_{\text{sel}}$):}
The selection cache $\mathbf{C}_{\text{sel}} \in \mathbb{R}^{C \times h \times w}$
stores the most recently \emph{retained} latent vector at each spatial location. Its sole purpose is to provide a temporal reference for token selection. At the beginning of a new chunk, the first latent frame is compared directly against $\mathbf{C}_{\text{sel}}$ rather than being automatically retained. If the temporal difference is below the threshold $\tau$, the token may be discarded; otherwise it is retained and the corresponding entry of $\mathbf{C}_{\text{sel}}$ is updated. This removes the artificial boundary effect present in the per-clip formulation.
\\
\item \textbf{Fill Cache ($\mathbf{C}_{\text{fill}}$):}
The fill cache $\mathbf{C}_{\text{fill}} \in \mathbb{R}^{C \times h \times w}$
stores the most recent \emph{non-zero latent presented to the inpainting network}. This cache is required because the first timestep of a chunk may now be discarded by the masking stage. After masking, discarded tokens are zeroed before being passed to the latent inpainting transformer (LIT). If the first timestep were removed without replacement, the inpainting network would receive no contextual latent information at that spatial location. Therefore, whenever a first-timestep token is dropped, its corresponding LIT input is populated using $\mathbf{C}_{\text{fill}}$.\\

\end{itemize}
The selection cache tracks the most recently \emph{kept} latent for temporal-difference computation, whereas the fill cache tracks the most recently \emph{available non-zero latent} used by the reconstruction pathway. Consequently, a single cache cannot simultaneously satisfy both requirements. Both caches operate independently at every spatial location and are reset prior to processing a new video to prevent latent-state leakage across sequences.

Table~\ref{tab:persistent-ref} compares the standard per-clip masking strategy against the proposed streaming variant on DAVIS at $\tau=0.30$. By exploiting temporal continuity across chunk boundaries, the cached approach substantially increases compression efficiency, reducing the keep rate from $62\%$ to $57\%$, while maintaining acceptable reconstruction fidelity. Pseudocode for the complete streaming algorithm is provided in \hyperref[sec:appendix-d]{Appendix~D} in the Supplementary material.

\begin{table}[!hbtp]
\centering
\small
\caption{
    Comparison of standard per-clip and streaming cached reference schemes on the DAVIS dataset at $\tau = 0.30$. The dual-cache variant carries latent states across chunk boundaries, eliminating periodic forced retention at clip starts and enabling more aggressive compression on long videos with continuous visual flow.
}
\vspace{0.2em}
\label{tab:persistent-ref}
\setlength{\tabcolsep}{6pt}
\begin{tabular}{@{}lccccc@{}}
\toprule
Variant & Keep Rate & PSNR$\uparrow$ & SSIM$\uparrow$ & LPIPS$\downarrow$ & FVD$\downarrow$ \\
\midrule
Per-clip reference & 62\% & 28.27 & 0.856 & 0.187 & 312.66 \\
Streaming cached & 57\% & 27.92 & 0.845 & 0.200 & 379.03 \\ 
\bottomrule
\end{tabular}
\vspace{-4mm}
\end{table}

\vspace{-2mm}
\section{Discussions and Limitations}
\label{sec:discussions}

\paragraph{Downstream task evaluation.}
Our evaluation focuses on reconstruction quality, but two questions about downstream use should be separated. The reconstructed latent $\hat{z}$ needs no adaptation by a consumer: it lies in the frozen backbone's native latent space and the $\mathcal{L}_{\text{latent}}$ term of Equation~\ref{eq:loss} supervises it towards $z$, making it a drop-in substitute wherever that backbone's latents are used. What remains open is whether the \emph{sparse retained set} can be consumed directly, without passing through LIT, which would require architectures accepting a variable number of positions per clip. We identify this as the primary direction for future work.

\paragraph{Temporal flicker}
Because the criterion drops different positions at different timesteps, the residual error varies over time and the visible artefact is flicker, to which per-frame metrics are blind. A pixel-space temporal-L1 term matching consecutive-frame differences reduces it in preliminary experiments; see \hyperref[sec:appendix-g]{Appendix~G} in the Supplementary material.

\paragraph{Limitation}
Because the compression rate emerges from the input rather than being specified upfront, our method cannot guarantee a target bits-per-pixel budget. On very high-motion content most positions exceed the threshold, few tokens are dropped, and little compression benefit remains. This is a property of the criterion, rate follows content, rather than a defect, but it limits applicability wherever a guaranteed rate is required.

\vspace{-2mm}
\section{Conclusion}
\label{sec:conclusion}

We presented a parameter-free approach to adaptive video tokenisation in the continuous-latent regime. Thresholding per-position temporal-L1 differences in a frozen tokeniser's latent space identifies redundant positions without any learned allocation network, and a factorised transformer (LIT) reconstructs the dropped positions from those retained, so inference costs a single encoder pass and no auxiliary decoder evaluation. Across TokenBench and DAVIS this is competitive with continuous-regime baselines at matched keep rates and lower cost.






\bibliography{egbib}

@article{agarwal2025cosmos,
  title           = {Cosmos World Foundation Model Platform for Physical {AI}},
  author          = {Agarwal, Niket and Ali, Arslan and Bala, Maciej and Balaji, Yogesh and Barker, Erik and Cai, Tiffany and Chattopadhyay, Prithvijit and Chen, Yongxin and Cui, Yin and Ding, Yifan and others},
  journal         = {arXiv preprint arXiv:2501.03575},
  year            = {2025}
}

@inproceedings{guo2024animatediffanimatepersonalizedtexttoimage,
  title           = {{A}nimate{D}iff: Animate Your Personalized Text-to-Image Diffusion Models without Specific Tuning}, 
  author          = {Guo, Yuwei and Yang, Ceyuan and Rao, Anyi and Liang, Zhengyang and Wang, Yaohui and Qiao, Yu and Agrawala, Maneesh and Lin, Dahua and Dai, Bo},
  booktitle       = {International Conference on Learning Representations (ICLR)},
  year            = {2024}
}

@inproceedings{omnitokenizer,
  title           = {{O}mni{T}okenizer: A Joint Image-Video Tokenizer for Visual Generation},
  author          = {Wang, Junke and Jiang, Yi and Yuan, Zehuan and Peng, Binyue and Wu, Zuxuan and Jiang, Yu-Gang},
  booktitle       = {Advances in Neural Information Processing Systems (NeurIPS)},
  year            = {2024}
}

@misc{liu2024sorareviewbackgroundtechnology,
  title           = {Sora: A Review on Background, Technology, Limitations, and Opportunities of Large Vision Models}, 
  author          = {Yixin Liu and Kai Zhang and Yuan Li and Zhiling Yan and Chujie Gao and Ruoxi Chen and Zhengqing Yuan and Yue Huang and Hanchi Sun and Jianfeng Gao and Lifang He and Lichao Sun},
  year            = {2024},
  eprint          = {2402.17177},
  archivePrefix   = {arXiv},
  primaryClass    = {cs.CV},
  url             = {https://arxiv.org/abs/2402.17177}, 
}

@inproceedings{Wang2024LARPTV,
  title           = {{LARP}: Tokenizing Videos with a Learned Autoregressive Generative Prior},
  author          = {Wang, Hanyu and Suri, Saksham and Ren, Yixuan and Chen, Hao and Shrivastava, Abhinav},
  booktitle       = {International Conference on Learning Representations (ICLR)},
  year            = {2025}
}

@inproceedings{magvit,
  title           = {{MAGVIT}: Masked generative video transformer},
  author          = {Yu, Lijun and Cheng, Yong and Sohn, Kihyuk and Lezama, Jos{\'e} and Zhang, Han and Chang, Huiwen and Hauptmann, Alexander G and Yang, Ming-Hsuan and Hao, Yuan and Essa, Irfan and Jiang, Lu},
  booktitle       = {Proceedings of the IEEE/CVF Conference on Computer Vision and Pattern Recognition (CVPR)},
  year            = {2023}
}

@inproceedings{adaptok,
  title           = {{A}dap{T}ok: Learning Adaptive and Temporally Causal Video Tokenization in a {1D} Latent Space},
  author          = {Li, Yan and Tian, Changyao and Xia, Renqiu and Liao, Ning and Guo, Weiwei and Li, Hongsheng and Dai, Jifeng and Li, Hao and Yang, Xue},
  booktitle       = {Proceedings of the IEEE/CVF Conference on Computer Vision and Pattern Recognition (CVPR)},
  year            = {2026}
}

@inproceedings{infotok,
  title           = {{I}nfo{T}ok: Adaptive Discrete Video Tokenizer via Information-Theoretic Compression},
  author          = {Ye, Haotian and He, Qiyuan and Han, Jiaqi and Li, Puheng and Fan, Jiaojiao and Hao, Zekun and Reda, Fitsum and Balaji, Yogesh and Chen, Huayu and Liu, Sheng and Yao, Angela and Zou, James and Ermon, Stefano and Wang, Haoxiang and Liu, Ming-Yu},
  booktitle       = {International Conference on Learning Representations (ICLR)},
  year            = {2026}
}

@inproceedings{Duggal2025SinglepassAI,
  title           = {Single-pass adaptive image tokenization for minimum program search},
  author          = {Shivam Duggal and Sanghyun Byun and William T. Freeman and Antonio Torralba and Phillip Isola},
  booktitle       = {Advances in Neural Information Processing Systems (NeurIPS)},
  year            = {2025}
}

@inproceedings{rlt,
  title           = {Don't Look Twice: Faster Video Transformers with Run-Length Tokenization},
  author          = {Choudhury, Rohan and Zhu, Guanglei and Liu, Sihan and Niinuma, Koichiro and Kitani, Kris M. and Jeni, L\'aszl\'o},
  booktitle       = {Advances in Neural Information Processing Systems (NeurIPS)},
  year            = {2024}
}

@article{videoworldsimulators2024,
  title           = {Video generation models as world simulators},
  author          = {Tim Brooks and Bill Peebles and Connor Holmes and Will DePue and Yufei Guo and Li Jing and David Schnurr and Joe Taylor and Troy Luhman and Eric Luhman and Clarence Ng and Ricky Wang and Aditya Ramesh},
  journal         = {OpenAI Research Technical Report},
  year            = {2024},
  url             = {https://openai.com/research/video-generation-models-as-world-simulators},
}

@article{assran2024learning,
  title           = {Revisiting feature prediction for learning visual representations from video},
  author          = {Bardes, Adrien and Garrido, Quentin and Ponce, Jean and Chen, Xinlei and Rabbat, Michael and LeCun, Yann and Assran, Mahmoud and Ballas, Nicolas},
  journal         = {arXiv preprint arXiv:2404.08471},
  year            = {2024}
}

@article{yan2021videogpt,
  title           = {{V}ideo{GPT}: Video Generation using {VQ-VAE} and Transformers},
  author          = {Wilson Yan and Yunzhi Zhang and Pieter Abbeel and Aravind Srinivas},
  journal         = {arXiv preprint arXiv:2104.10157},
  year            = {2021}
}

@inproceedings{zhao2024cvvaecompatiblevideovae,
  title           = {{CV-VAE}: A Compatible Video {VAE} for Latent Generative Video Models}, 
  author          = {Sijie Zhao and Yong Zhang and Xiaodong Cun and Shaoshu Yang and Muyao Niu and Xiaoyu Li and Wenbo Hu and Ying Shan},
  booktitle       = {Advances in Neural Information Processing Systems (NeurIPS)},
  year            = {2024}
}

@inproceedings{blattmann2023alignlatentshighresolutionvideo,
  title           = {Align your Latents: High-Resolution Video Synthesis with Latent Diffusion Models},
  author          = {Blattmann, Andreas and Rombach, Robin and Ling, Huan and Dockhorn, Tim and Kim, Seung Wook and Fidler, Sanja and Kreis, Karsten},
  booktitle       = {IEEE Conference on Computer Vision and Pattern Recognition ({CVPR})},
  year            = {2023}
}

@inproceedings{yu2024language,
  title           = {Language model beats diffusion -- tokenizer is key to visual generation},
  author          = {Yu, Lijun and Lezama, Jos{\'e} and Gundavarapu, Nitesh Bharadwaj and Versari, Luca and Sohn, Kihyuk and Minnen, David and Cheng, Yong and Gupta, Agrim and Gu, Xiuye and Hauptmann, Alexander G and others},
  booktitle       = {International Conference on Learning Representations (ICLR)},
  year            = {2024}
}

@misc{nvidia2024cosmos,
  title           = {Cosmos Tokenizer: High-Fidelity Video and Image Compression for World Models},
  author          = {{NVIDIA Cosmos Tokenizer Team}},
  howpublished    = {GitHub Repository},
  year            = {2024},
  url             = {https://github.com/NVIDIA/Cosmos-Tokenizer}
}

@inproceedings{vaswani2017attention,
  title           = {Attention is All You Need},
  author          = {Vaswani, Ashish and Shazeer, Noam and Parmar, Niki and Uszkoreit, Jakob and Jones, Llion and Gomez, Aidan N and Kaiser, {\L}ukasz and Polosukhin, Illia},
  booktitle       = {Advances in Neural Information Processing Systems (NeurIPS)},
  year            = {2017}
}

@misc{caelles20192019davischallengevos,
  title           = {The 2019 {DAVIS} Challenge on {VOS}: Unsupervised Multi-Object Segmentation}, 
  author          = {Sergi Caelles and Jordi Pont-Tuset and Federico Perazzi and Alberto Montes and Kevis-Kokitsi Maninis and Luc Van Gool},
  year            = {2019},
  eprint          = {1905.00737},
  archivePrefix   = {arXiv},
  primaryClass    = {cs.CV},
  url             = {https://arxiv.org/abs/1905.00737}, 
}

@inproceedings{dosovitskiy2020image,
  title           = {An Image is Worth 16x16 Words: Transformers for Image Recognition at Scale},
  author          = {Dosovitskiy, Alexey and Beyer, Lucas and Kolesnikov, Alexander and Weissenborn, Dirk and Zhai, Xiaohua and Unterthiner, Thomas and Dehghani, Mostafa and Minderer, Matthias and Heigold, Georg and Gelly, Sylvain and others},
  booktitle       = {International Conference on Learning Representations (ICLR)},
  year            = {2021}
}

@inproceedings{zhang2018unreasonableeffectivenessdeepfeatures,
  title           = {The Unreasonable Effectiveness of Deep Features as a Perceptual Metric}, 
  author          = {Zhang, Richard and Isola, Phillip and Efros, Alexei A. and Shechtman, Eli and Wang, Oliver},
  booktitle       = {Proceedings of the IEEE/CVF Conference on Computer Vision and Pattern Recognition (CVPR)},
  year            = {2018}
}

@misc{unterthiner2019accurategenerativemodelsvideo,
  title           = {Towards Accurate Generative Models of Video: A New Metric \& Challenges}, 
  author          = {Thomas Unterthiner and Sjoerd van Steenkiste and Karol Kurach and Raphael Marinier and Marcin Michalski and Sylvain Gelly},
  year            = {2019},
  eprint          = {1812.01717},
  archivePrefix   = {arXiv},
  primaryClass    = {cs.CV},
  url             = {https://arxiv.org/abs/1812.01717}, 
}

@inproceedings{arnab2021vivit,
  title           = {{ViViT}: A Video Vision Transformer},
  author          = {Arnab, Anurag and Dehghani, Mostafa and Heigold, Georg and Sun, Chen and Lu{\v{c}}i{\'c}, Mario and Schmid, Cordelia},
  booktitle       = {Proceedings of the IEEE/CVF International Conference on Computer Vision (ICCV)},
  year            = {2021}
}

@inproceedings{yan2025elastictok,
  title           = {{E}lastic{T}ok: Adaptive tokenization for image and video},
  author          = {Yan, Wilson and Mnih, Volodymyr and Faust, Aleksandra and Zaharia, Matei and Abbeel, Pieter and Liu, Hao},
  booktitle       = {International Conference on Learning Representations (ICLR)},
  year            = {2025}
}

@article{su2023roformerenhancedtransformerrotary,
  title           = {{R}o{F}ormer: Enhanced Transformer with Rotary Position Embedding}, 
  author          = {Jianlin Su and Murtadha Ahmed and Yu Lu and Shengfeng Pan and Wen Bo and Yunfeng Liu},
  journal         = {Neurocomputing},
  year            = {2024}
}

@inproceedings{esser2021taming,
  title           = {Taming transformers for high-resolution image synthesis},
  author          = {Esser, Patrick and Rombach, Robin and Ommer, Bjorn},
  booktitle       = {Proceedings of the IEEE/CVF Conference on Computer Vision and Pattern Recognition (CVPR)},
  year            = {2021}
}

@inproceedings{loshchilov2019decoupledweightdecayregularization,
  title           = {Decoupled Weight Decay Regularization},
  author          = {Ilya Loshchilov and Frank Hutter},
  booktitle       = {International Conference on Learning Representations (ICLR)},
  year            = {2019}
}

@inproceedings{mondal2020maskaaelatentspaceoptimization,
  title           = {{M}ask{AAE}: Latent space optimization for Adversarial Auto-Encoders}, 
  author          = {Arnab Kumar Mondal and Sankalan Pal Chowdhury and Aravind Jayendran and Parag Singla and Himanshu Asnani and Prathosh AP},
  booktitle       = {Conference on Uncertainty in Artificial Intelligence (UAI)},
  year            = {2020}
}

@misc{hu2024matryoshkaquerytransformerlarge,
  title           = {Matryoshka Query Transformer for Large Vision-Language Models}, 
  author          = {Wenbo Hu and Zi-Yi Dou and Liunian Harold Li and Amita Kamath and Nanyun Peng and Kai-Wei Chang},
  year            = {2024},
  eprint          = {2405.19315},
  archivePrefix   = {arXiv},
  primaryClass    = {cs.CV},
  url             = {https://arxiv.org/abs/2405.19315}, 
}

@article{kay2017kinetics,
  title           = {The {K}inetics human action video dataset},
  author          = {Kay, Will and Carreira, Joao and Simonyan, Karen and Zhang, Brian and Hillier, Chloe and Vijayanarasimhan, Sudheendra and Viola, Fabio and Green, Tim and Back, Trevor and Natsev, Paul and others},
  journal         = {arXiv preprint arXiv:1705.06950},
  year            = {2017}
}

@inproceedings{van2017neural,
  title           = {Neural discrete representation learning},
  author          = {van den Oord, Aaron and Vinyals, Oriol and Kavukcuoglu, Koray},
  booktitle         = {Advances in Neural Information Processing Systems (NeurIPS)},
  year            = {2017}
}

@inproceedings{deng2025autoregressivevideogenerationvector,
  title           = {Autoregressive Video Generation without Vector Quantization}, 
  author          = {Deng, Haoge and Pan, Ting and Diao, Haiwen and Luo, Zhengxiong and Cui, Yufeng and Lu, Huchuan and Shan, Shiguang and Qi, Yonggang and Wang, Xinlong},
  booktitle       = {International Conference on Learning Representations (ICLR)},
  year            = {2025}
}

@misc{soomro2012ucf101dataset101human,
  title           = {{UCF101}: A Dataset of 101 Human Actions Classes From Videos in The Wild}, 
  author          = {Khurram Soomro and Amir Roshan Zamir and Mubarak Shah},
  year            = {2012},
  eprint          = {1212.0402},
  archivePrefix   = {arXiv},
  primaryClass    = {cs.CV},
  url             = {https://arxiv.org/abs/1212.0402}, 
}

@article{ssim,
  title           = {Image quality assessment: from error visibility to structural similarity},
  author          = {Zhou Wang and Bovik, A.C. and Sheikh, H.R. and Simoncelli, E.P.},
  journal         = {IEEE Transactions on Image Processing},
  year            = {2004}
}

\clearpage

\begin{center}
    {\Large\bf Supplementary Material}\\[0.4em]
    {\large Adaptive Tokenisation Via Temporal Redundancy Masking And Latent Inpainting.}
\end{center}
\vspace{1em}

\raggedbottom

\phantomsection
\section*{Appendix A: Empirical Threshold Calibration}
\label{sec:appendix-a}

Our mechanism relies on a fixed scalar threshold, $\tau$, which dictates whether the temporal L1 difference $\Delta(t', y, x)$ between consecutive latent positions is perceptually relevant.

While this philosophy is conceptually similar to the data-agnostic thresholding proposed in Run-Length Tokenisation (RLT), the continuous latent regime requires backbone-specific calibration. RLT operates directly in normalised pixel space, whereas our method operates within the continuous manifolds of pre-trained video tokenisers. Because foundational tokenisers like Cosmos and OmniTokenizer enforce different absolute scales and variances upon their latent spaces, $\tau$ must be calibrated to the specific backbone.

To determine the nominal operating point, we conducted an empirical sweep over a range of candidate $\tau$ values. For each candidate threshold, we generated the corresponding temporal-L1 masks across a held-out subset of videos and evaluated the resulting global keep rates alongside visual reconstruction quality. As $\tau$ increases, the criterion becomes more aggressive, dropping more tokens but risking the loss of subtle motion details.

Through this empirical evaluation, we determined that $\tau = 0.3$ is the optimal operating point for the Cosmos backbone. At this threshold, the generated masks reliably identify and drop static spatial-temporal redundancies while strictly preserving the boundaries of genuine motion, yielding an average keep rate of roughly 48\% on UCF-101. An identical empirical sweep performed on the OmniTokenizer backbone—which exhibits a wider intrinsic latent variance—led to the selection of $\tau = 1.2$.

\begin{figure}[H]
    \centering
    \includegraphics[width=\linewidth]{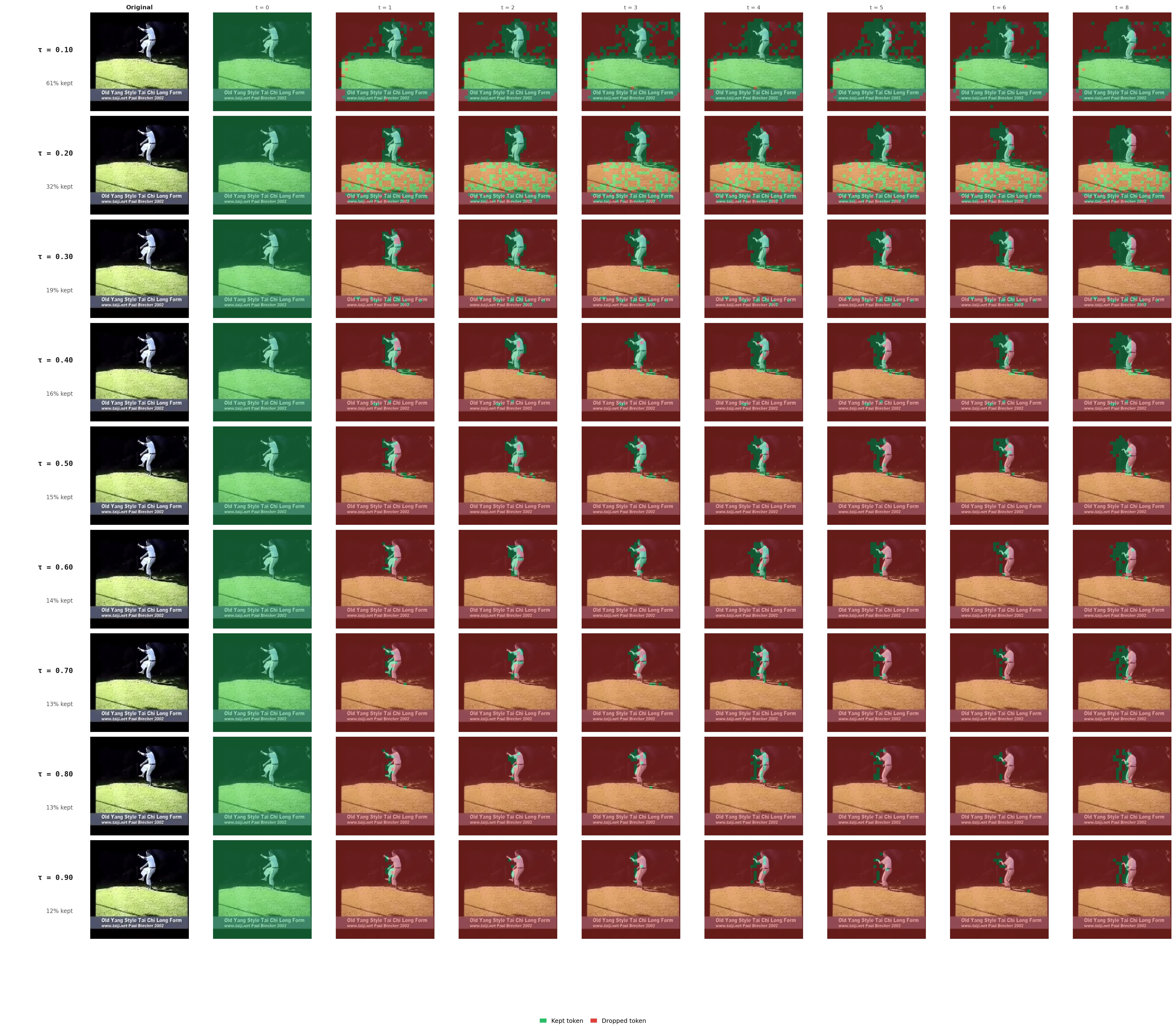}
    \vspace{0.5em}
    \caption{
        \textbf{Empirical calibration of the threshold $\tau$.}
         Visualisation of the generated temporal masks at varying $\tau$ values (e.g., $\tau \in \{0.1, 0.2, 0.3, 0.4, \dots, 0.9\}$). By empirically assessing these masks, we established $\tau = 0.3$ as the optimal balance for the cosmos backbone, aggressively compressing static backgrounds without compromising high-frequency dynamic regions.
    }
    \label{fig:supp_calibration}
\end{figure}

\begin{figure}[H]
    \centering
    \includegraphics[width=\linewidth]{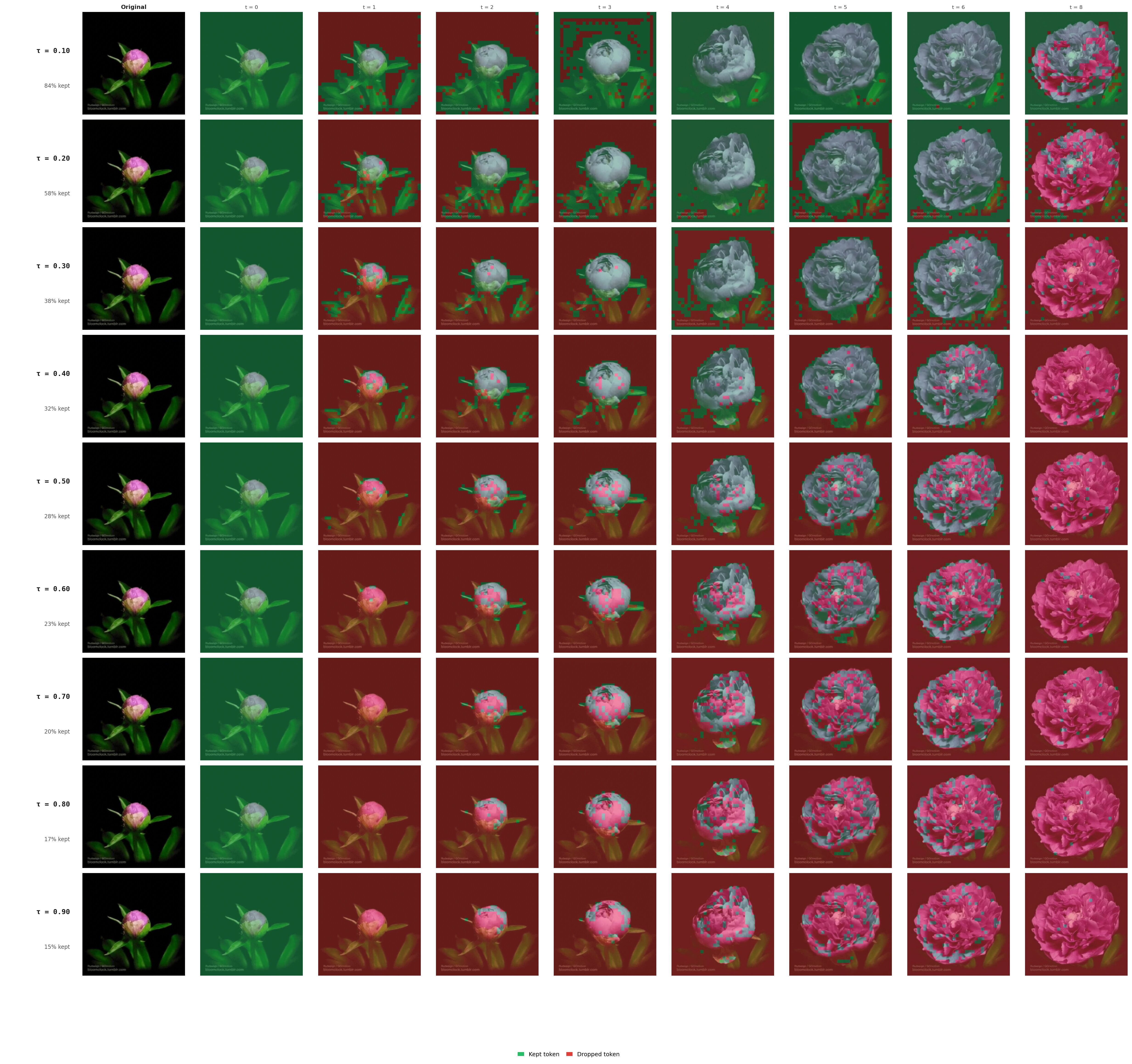}
    \caption{Additional example for our empirical calibration of the threshold $\tau$ for Cosmos Backbone.}
    \label{fig:supp_tau_cosmos_flower}
\end{figure}

\begin{figure}[H]
    \centering
    \includegraphics[width=\linewidth]{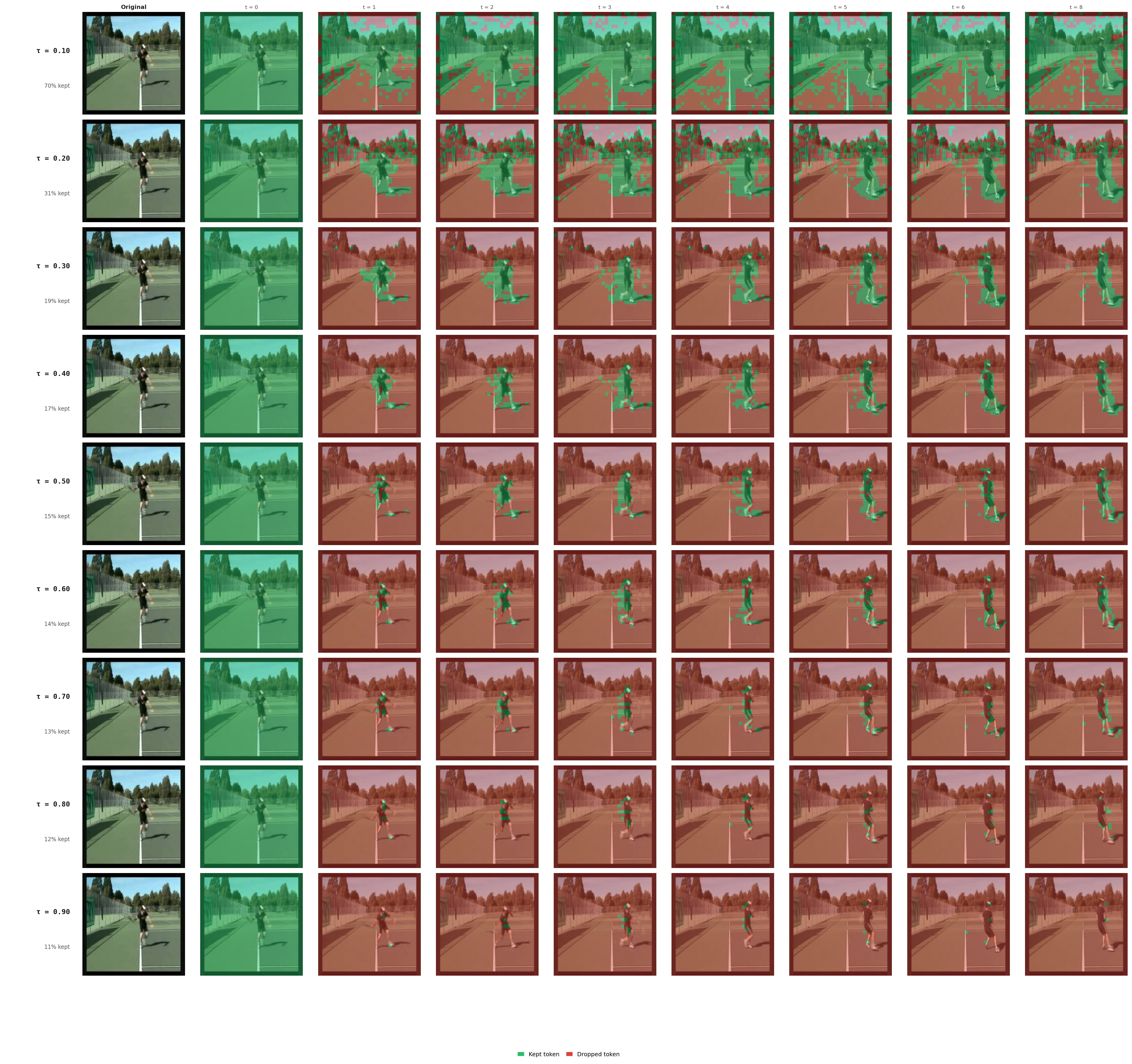}
    \caption{Additional example for our empirical calibration of the threshold $\tau$ for Cosmos Backbone.}
    \label{fig:supp_tau_cosmos_tennis}
\end{figure}

\begin{figure}[H]
    \centering
    \includegraphics[width=\linewidth]{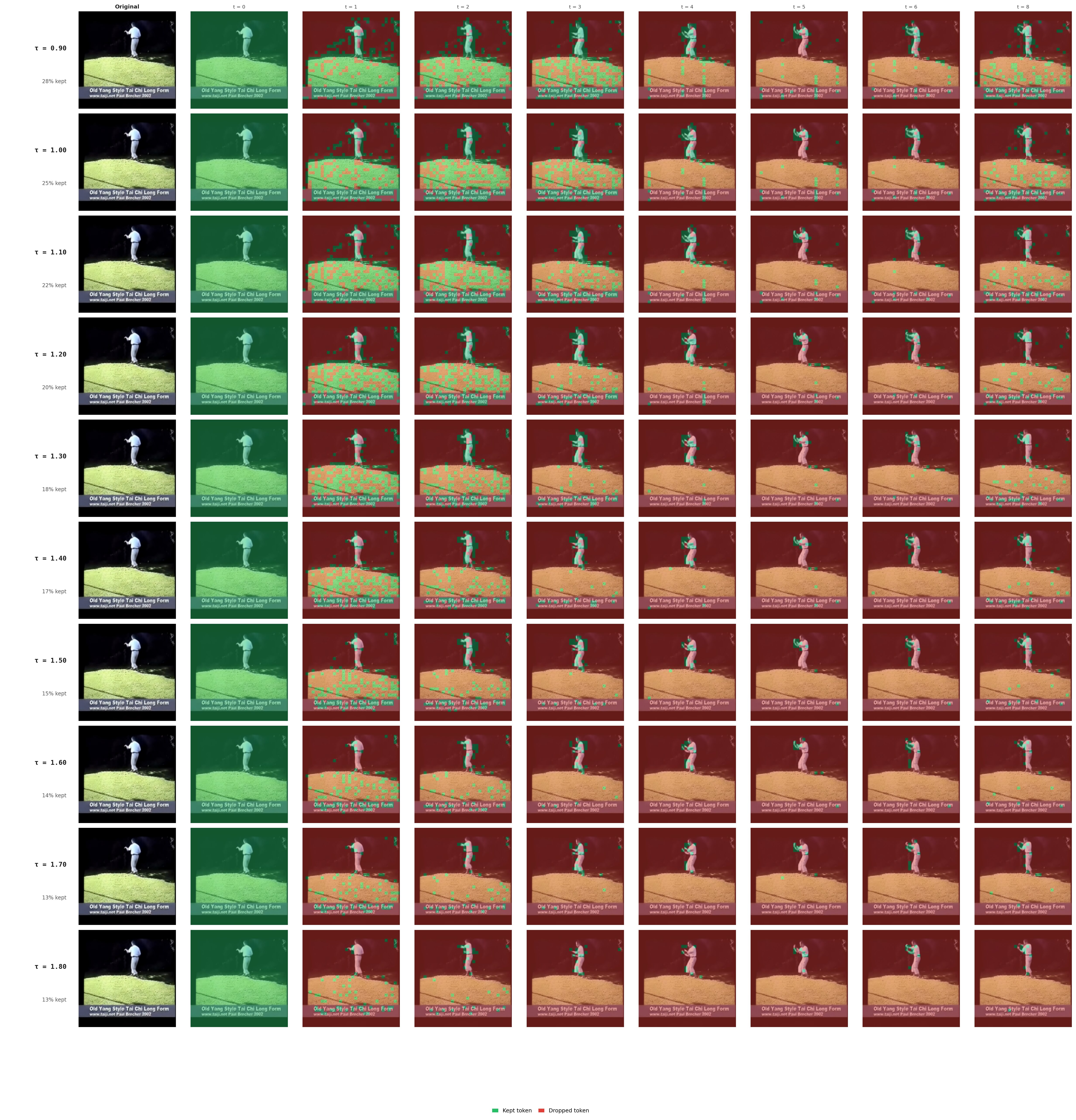}
    \vspace{0.5em}
    \caption{
        \textbf{Empirical calibration of the threshold $\tau$.}
         Visualisation of the generated temporal masks at varying $\tau$ values (e.g., $\tau \in \{0.9, 1.0, 1.1, 1.2, \dots, 1.8\}$). By empirically assessing these masks, we established $\tau = 1.2$ as the optimal balance for the omni backbone.
    }
    \label{fig:supp_calibration_omni}
\end{figure}

\begin{figure}[H]
    \centering
    \includegraphics[width=\linewidth]{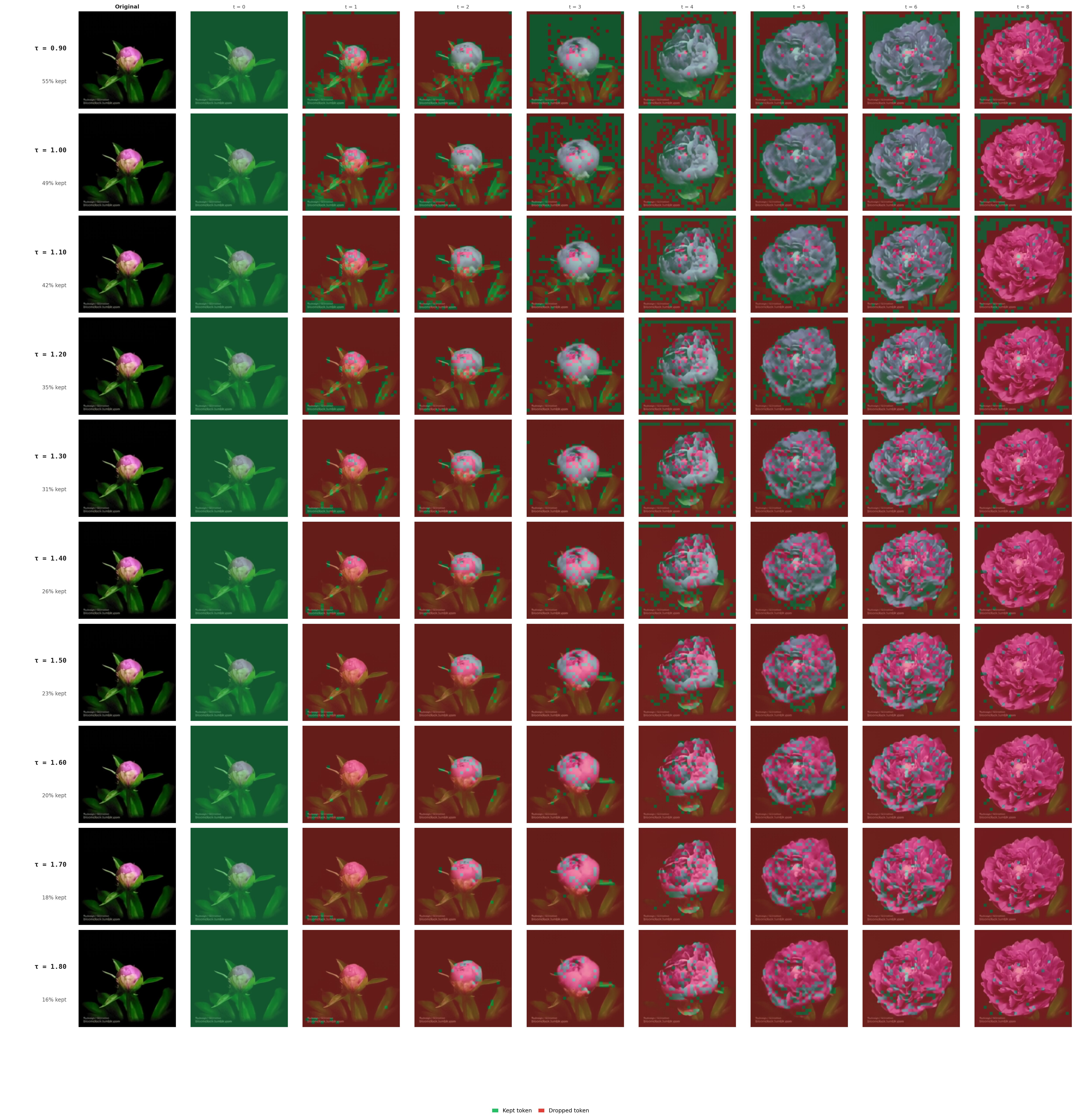}
    \caption{Additional example for our empirical calibration of the threshold $\tau$ for Omni Backbone.}
    \label{fig:supp_tau_omni_flower}
\end{figure}

\begin{figure}[H]
    \centering
    \includegraphics[width=\linewidth]{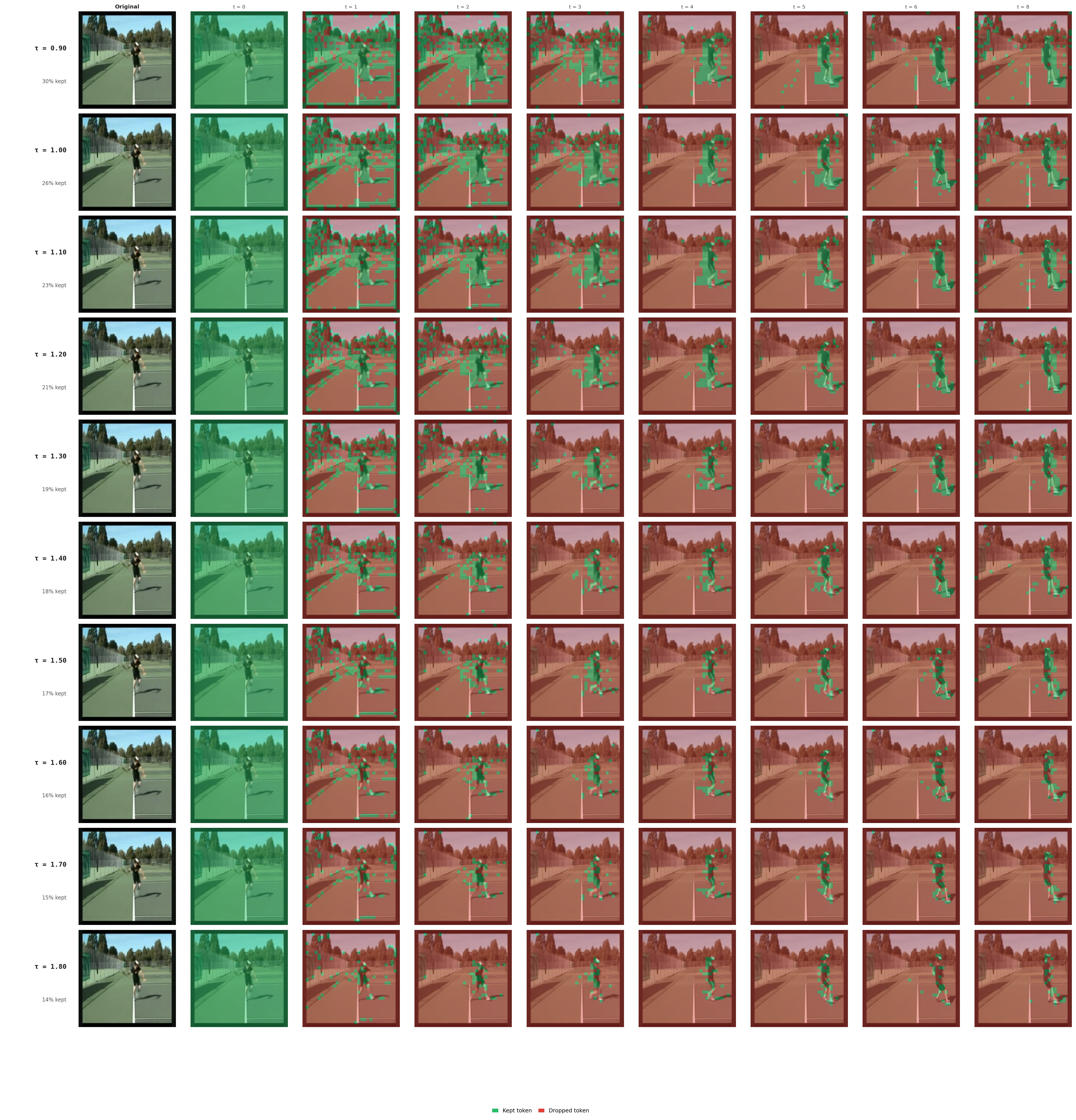}
    \caption{Additional example for our empirical calibration of the threshold $\tau$ for Omni Backbone.}
    \label{fig:supp_tau_omni_tennis}
\end{figure}

\phantomsection
\subsection*{A.1: Threshold Sensitivity}
\label{sec:appendix-a-sensitivity}

\begin{figure}[H]
    \centering
    \includegraphics[width=\linewidth]{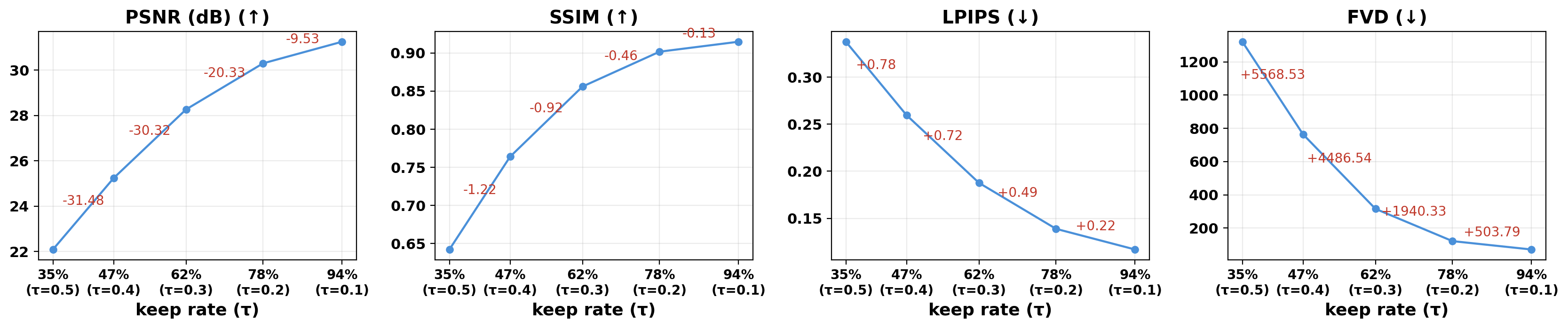}
    \caption{\textbf{Cosmos--DAVIS $\tau$ sweep.} Reconstruction quality against the emergent keep rate, with the threshold that produced each operating point annotated beneath it. Quality varies smoothly and monotonically over the whole range $\tau \in [0.1, 0.5]$, spanning keep rates from $94\%$ down to $35\%$.}
    \label{fig:sweep-cosmos-davis}
\end{figure}

\begin{figure}[H]
    \centering
    \includegraphics[width=\linewidth]{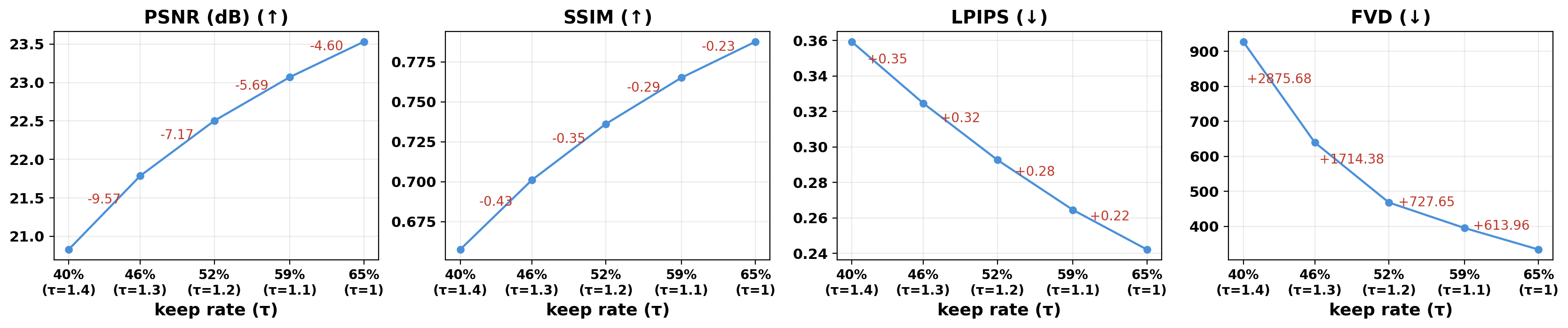}
    \caption{\textbf{OmniTokenizer--DAVIS $\tau$ sweep.} The same sweep with $\tau$ re-tuned to OmniTokenizer's latent scale, $\tau \in [1.0, 1.4]$ against $[0.1, 0.5]$ for Cosmos (\hyperref[sec:appendix-e]{Appendix~E}). The curve is likewise monotonic, confirming that the criterion transfers across backbones once the threshold is placed on the correct scale.}
    \label{fig:sweep-omni-davis}
\end{figure}

The visual sweeps above establish the operating point qualitatively. To characterise the criterion quantitatively across its full range, we sweep $\tau$ on the trained model and record the resulting keep rate together with reconstruction quality. The same trained LIT is used at every point; no retraining is performed. Figures~\ref{fig:sweep-cosmos-davis} and~\ref{fig:sweep-omni-davis} plot PSNR, SSIM, LPIPS and FVD against the emergent fraction of tokens kept.

Two properties are evident. First, quality degrades \emph{smoothly and monotonically} as the keep rate falls, across all four metrics and both backbones. There is no threshold at which reconstruction collapses, and no narrow basin of good behaviour that would have to be located by search. This is what makes a single calibrated $\tau$ a workable operating point rather than a fragile one, and it is also what allows a single trained LIT to be deployed at a range of quality--rate trade-offs without retraining. Second, the Cosmos sweep, which extends to $94\%$ retention, flattens at its upper end: beyond roughly $80\%$ the marginal quality returned per additional token is small, which is precisely the redundancy the method is designed to exploit. The OmniTokenizer sweep covers a narrower band ($40$--$65\%$) and is still climbing at its upper end, consistent with the lower masking-score selectivity analysed in \hyperref[sec:appendix-e]{Appendix~E}.

\phantomsection
\section*{Appendix B: Mask Visualisation}
\label{sec:appendix-b}

\begin{figure}[H]
    \centering

    \includegraphics[width=\linewidth]{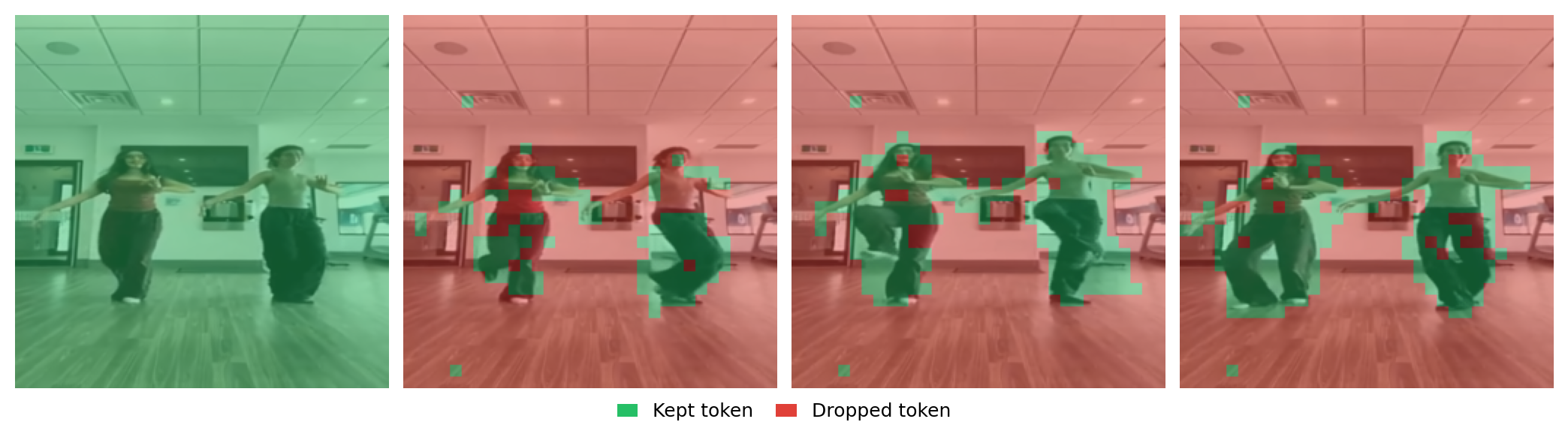}

    \vspace{0.5em}

    \includegraphics[width=\linewidth]{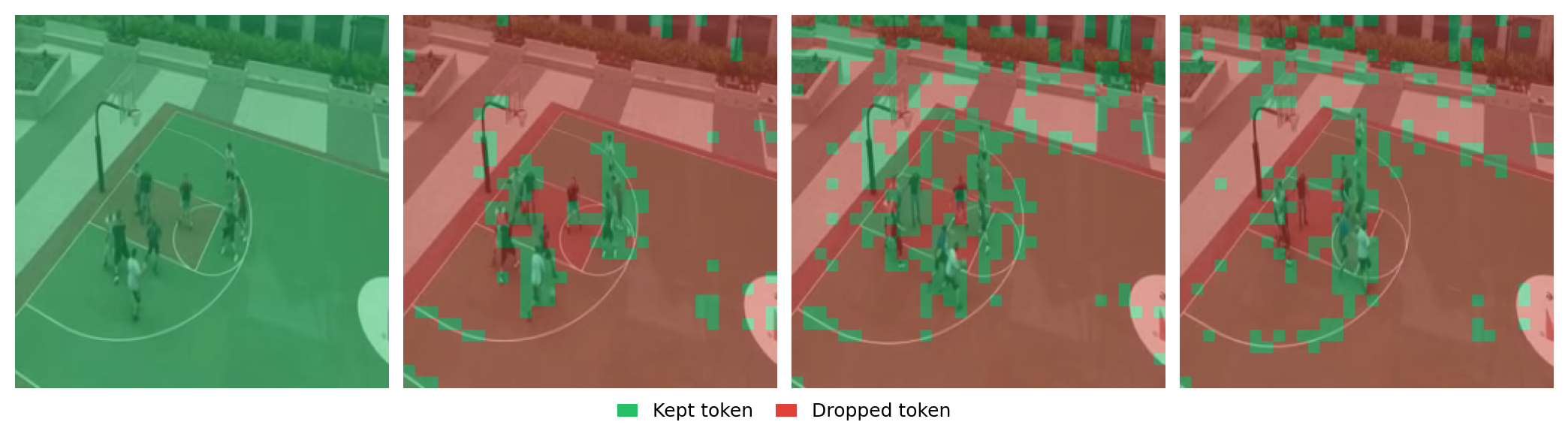}

    \vspace{0.5em}

    \includegraphics[width=\linewidth]{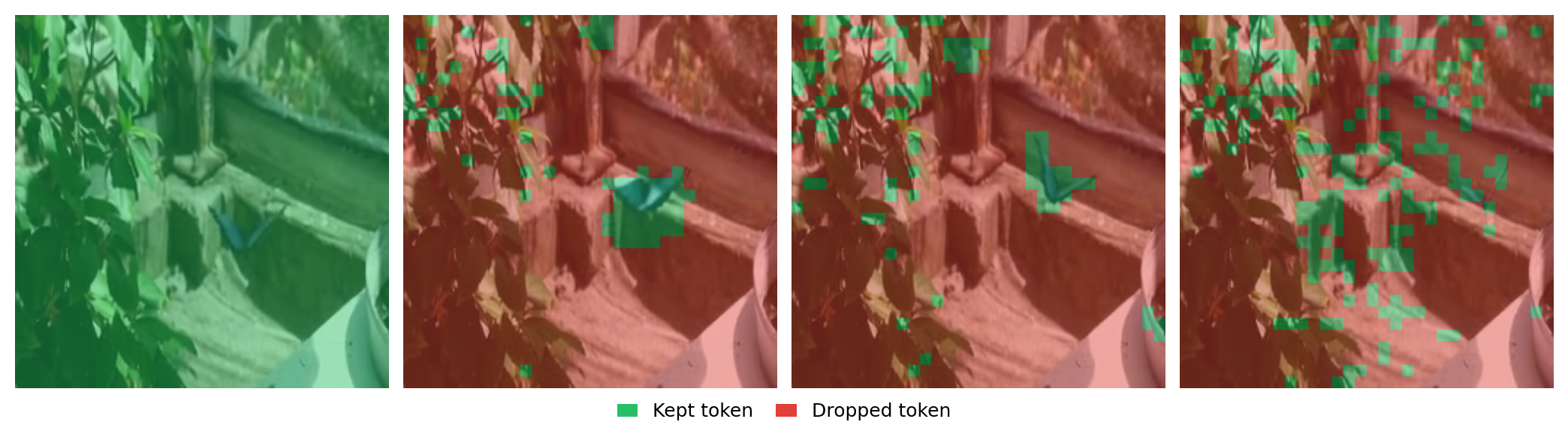}

    \vspace{0.5em}

    \includegraphics[width=\linewidth]{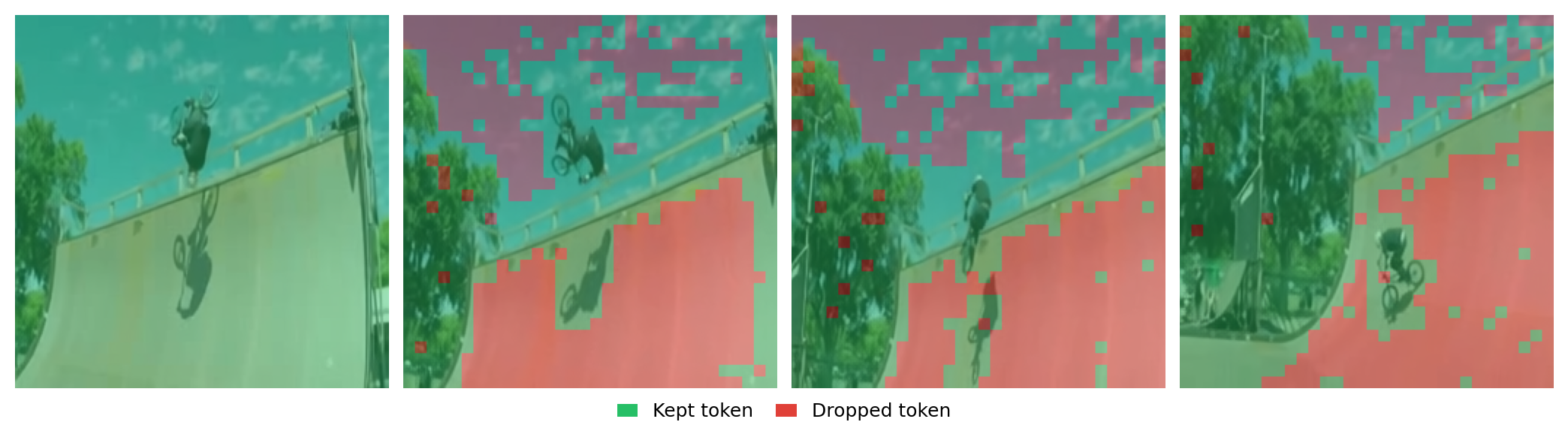}

    \caption{
        Temporal masks on representative videos.
        First: Masks on a video of people dancing.
        Second: Masks on a video of people playing basketball.
        Third: Masks on a video of butterfly fluttering.
        Fourth: Masks on a video of person performing BMX stunts.
    }

    \label{fig:mask_examples}
\end{figure}

\phantomsection
\section*{Appendix C: Token Utilisation Across Chunk Boundaries}
\label{sec:appendix-c}

\begin{figure}[H]
    \centering
    \includegraphics[width=0.75\linewidth]{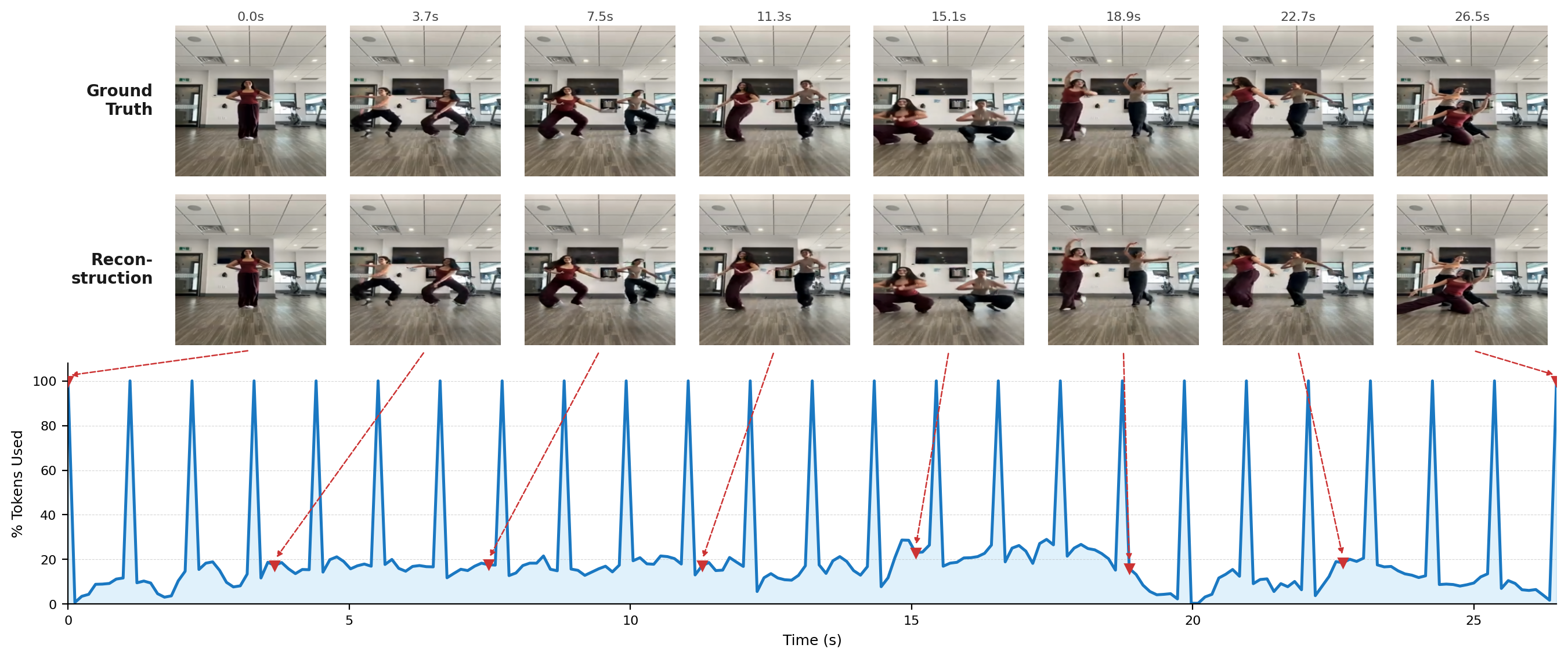}
    \vspace{0.5em}
    \includegraphics[width=0.75\linewidth]{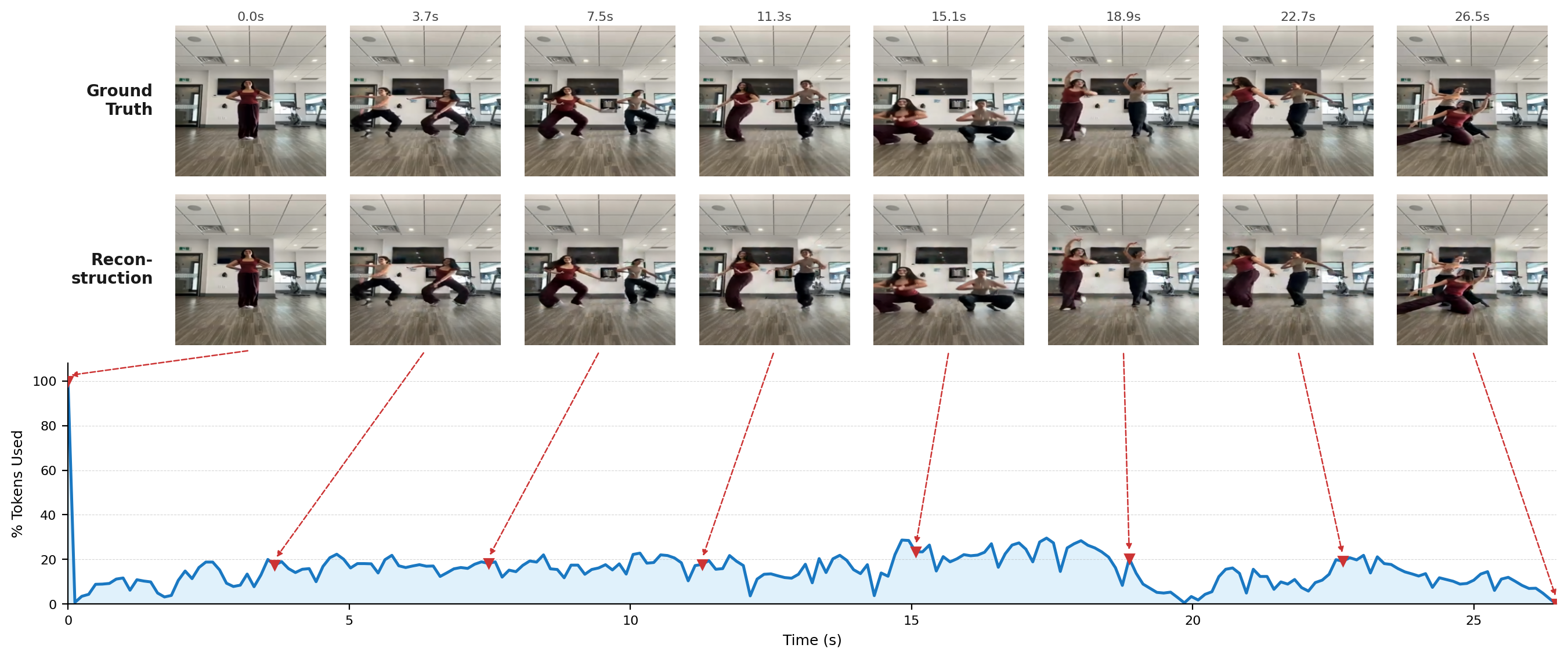}
    \caption{\textbf{Token utilisation over time, per-clip against streaming.} Percentage of latent tokens retained across a $27$-second sequence at $\tau = 0.30$. \textbf{Top:} the per-clip reference regime, where the reference resets at every $33$-frame chunk boundary and forces full retention roughly once per second. \textbf{Bottom:} the streaming cached regime, where the selection and fill caches persist across boundaries and only the first frame of the video is retained in full. Reconstruction quality is unchanged.
    }
    \label{fig:supp_token_usage}
\end{figure}

Figure~\ref{fig:supp_token_usage} makes the chunk-boundary effect that motivates the streaming variant directly visible. Both panels track the percentage of latent tokens retained over time for the same $27$-second dance sequence at $\tau = 0.30$, with ground-truth and reconstructed frames shown above each curve.

Under the per-clip reference (top), retention returns to $100\%$ once per $33$-frame chunk, roughly every $1.1$ seconds here. The last-kept reference is reinitialised at every boundary, so the first latent frame of each chunk is retained in full regardless of how little has changed since the end of the previous chunk. Between spikes the criterion behaves as intended, holding retention in the $10$--$25\%$ band and rising only where the dancers move quickly. The spikes are therefore an artefact of clip-wise processing rather than of video content: they encode nothing the previous chunk had not already encoded.

Under the streaming dual-cache reference (bottom) they are gone. Only the first frame of the video is retained in full; thereafter retention follows scene motion continuously, with no periodic structure. The saving is precisely the area under those spikes, and aggregated over the dataset it is what reduces the keep rate from $62\%$ to $57\%$ in Table~\ref{tab:persistent-ref}. The reconstructions above the two curves are visually indistinguishable, confirming that the tokens removed at the boundaries were redundant.

\clearpage
\phantomsection
\section*{Appendix D: Streaming Cached Temporal Masking}
\label{sec:appendix-d}

The streaming subsection of the main paper motivates the dual-cache variant and reports its results. Algorithm~\ref{alg:streaming_cached} states the complete procedure.

\begin{algorithm}[H]
\caption{Streaming Cached Temporal Masking (Dual-Cache)}
\label{alg:streaming_cached}
\begin{algorithmic}[1]
    \Require Video chunks $\{\mathbf{X}^{(1)}, \dots, \mathbf{X}^{(K)}\}$, Encoder $\mathcal{E}$, LIT parameters $\theta$, Frozen decoder $\mathcal{D}$, Threshold $\tau$
    \Ensure Reconstructed video chunks $\{\mathbf{Y}^{(1)}, \dots, \mathbf{Y}^{(K)}\}$, Masks $\{\mathbf{M}^{(1)}, \dots, \mathbf{M}^{(K)}\}$
    
    \State $\mathbf{C}_{\text{sel}} \gets \mathbf{0}, \quad \mathbf{C}_{\text{fill}} \gets \mathbf{0}$
    \State $\text{isFirstChunk} \gets \text{True}$
    
    \For{$k = 1$ to $K$}
        \State $\mathbf{Z}^{(k)} \gets \mathcal{E}(\mathbf{X}^{(k)})$ 
        \State Initialize mask $\mathbf{M}^{(k)} \gets \mathbf{0}$
        
        \For{$t = 1$ to $T$}
            \If{$t = 1$ \textbf{and} $\text{isFirstChunk}$}
                \State $\mathbf{M}_t^{(k)} \gets \mathbf{1}$ \Comment{Always retain first frame of the first chunk}
            \Else
                \State $\Delta_t \gets \frac{1}{C} \sum_{c=1}^C \bigl| \mathbf{Z}_t^{(k)} - \mathbf{C}_{\text{sel}} \bigr|$
                \State $\mathbf{M}_t^{(k)} \gets \mathds{1}[\Delta_t \geq \tau]$
            \EndIf
            
            \State $\mathbf{C}_{\text{sel}} \gets \text{where}(\mathbf{M}_t^{(k)} = \mathbf{1}, \, \mathbf{Z}_t^{(k)}, \, \mathbf{C}_{\text{sel}})$ \Comment{Update selection cache}
        \EndFor

        \State
        
        \State $\mathbf{Z}_{\text{masked}}^{(k)} \gets \mathbf{Z}^{(k)} \odot \mathbf{M}^{(k)}$
        \State
        \State $\mathbf{Z}_{\text{in}}^{(k)} \gets \mathbf{Z}_{\text{masked}}^{(k)}$
        
        \State $\mathbf{Z}_{\text{in}, 1}^{(k)} \gets \text{where}(\mathbf{M}_1^{(k)} = \mathbf{1}, \, \mathbf{Z}_{\text{in}, 1}^{(k)}, \, \mathbf{C}_{\text{fill}})$ \Comment{Fill dropped boundary tokens using Fill Cache}
        
        \State
        \For{$t = 1$ to $T$}
            \State $\mathbf{C}_{\text{fill}} \gets \text{where}(\mathbf{Z}_{\text{in}, t}^{(k)} \neq \mathbf{0}, \, \mathbf{Z}_{\text{in}, t}^{(k)}, \, \mathbf{C}_{\text{fill}})$ \Comment{Update Fill Cache}
        \EndFor
        
        \State $\mathbf{Z}_{\text{rec}}^{(k)} \gets \text{LIT}_{\theta}(\mathbf{Z}_{\text{in}}^{(k)})$ \Comment{Inpaint dropped positions}
        \State $\mathbf{Y}^{(k)} \gets \mathcal{D}(\mathbf{Z}_{\text{rec}}^{(k)})$ \Comment{Decode with frozen decoder}
        
        \State $\text{isFirstChunk} \gets \text{False}$
    \EndFor
    
    \State \Return $\{\mathbf{Y}^{(k)}\}_{k=1}^K, \{\mathbf{M}^{(k)}\}_{k=1}^K$
\end{algorithmic}
\end{algorithm}

\phantomsection
\section*{Appendix E: OmniTokenizer Backbone Analysis}
\label{sec:appendix-e}

Our method is backbone-agnostic in formulation: it requires only a continuous latent grid and a scalar threshold. In practice, however, the absolute margins reported in the main paper are noticeably smaller on the OmniTokenizer backbone than on Cosmos. Since the masking criterion itself is unchanged between the two, the difference must originate in the latent spaces they induce. We analyse this along two axes: the \emph{scale} of the latent space, which determines where $\tau$ must be placed, and the \emph{selectivity} of the resulting masking scores, which determines how cleanly any single threshold can separate redundant from informative positions.

\begin{figure}[H]
    \centering
    \includegraphics[width=0.78\linewidth]{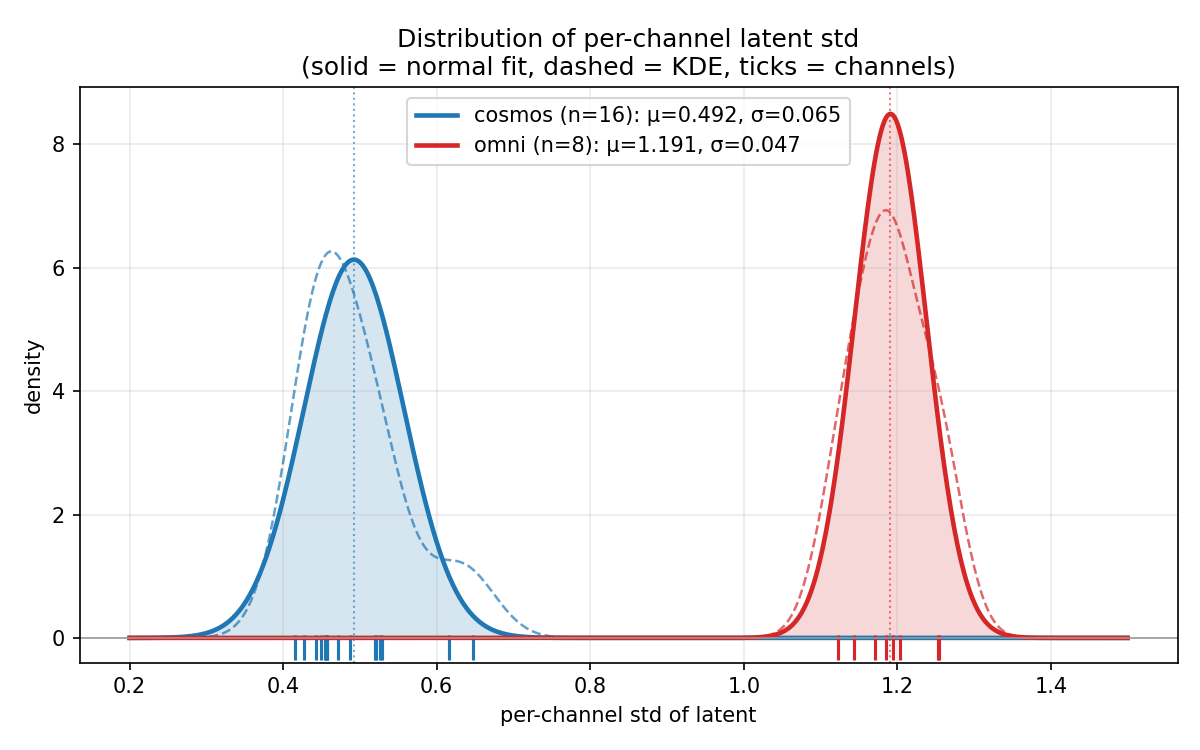}

    \vspace{0.6em}

    \includegraphics[width=0.78\linewidth]{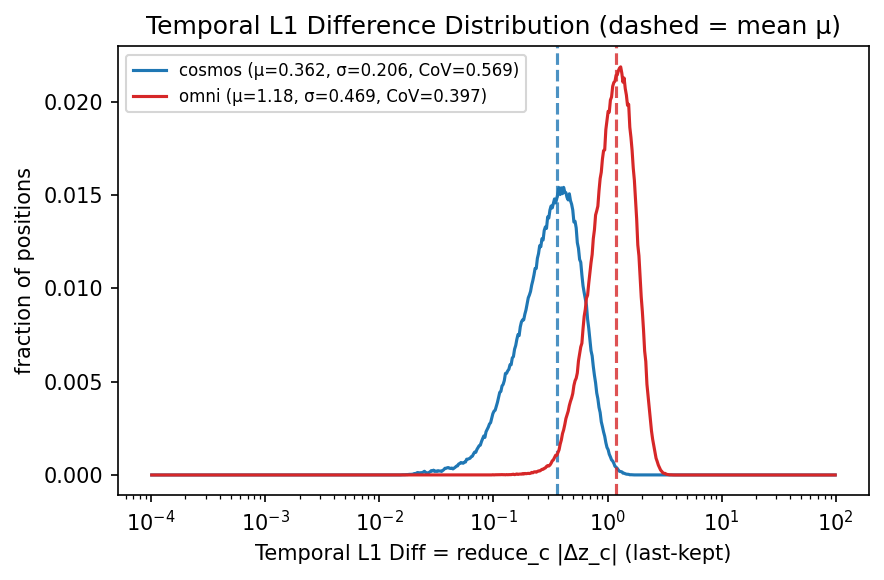}
    \caption{
        \textbf{Why OmniTokenizer differs from Cosmos.}
        Top: distribution of per-channel latent standard deviations. OmniTokenizer is centred at
        $\mu = 1.191$ against $0.492$ for Cosmos, and carries half as many channels ($8$ against $16$).
        Bottom: distribution of temporal-L1 masking scores under the last-kept reference. OmniTokenizer
        sits at $\mu = 1.18$ ($\sigma = 0.469$, CoV $= 0.397$) and Cosmos at $\mu = 0.362$
        ($\sigma = 0.206$, CoV $= 0.569$). The calibrated threshold for each backbone,
        $\tau = 1.2$ and $\tau = 0.3$ respectively, falls close to the mean of its own score
        distribution.
    }
    \label{fig:omni-analysis}
\end{figure}

\paragraph{Scale.}
Our criterion thresholds an \emph{absolute} L1 difference (the masking criterion of the main paper), so the numerical value of the masking score inherits the magnitude of the latent space it is computed in. The two backbones differ substantially in this respect (Figure~\ref{fig:omni-analysis}): the distribution of per-channel latent standard deviations is centred at $\mu = 1.191$ for OmniTokenizer against $\mu = 0.492$ for Cosmos, roughly $2.4\times$ larger, and OmniTokenizer carries half as many channels ($8$ against $16$), so the channel-mean of Equation~1 averages over fewer samples and is correspondingly noisier. The masking scores inherit this directly, sitting at $\mu = 1.18$ for OmniTokenizer against $\mu = 0.362$ for Cosmos. A threshold calibrated on Cosmos is therefore far too permissive when applied unchanged to OmniTokenizer, since the same $\tau$ occupies a much lower quantile of the score distribution. This is not a defect of the criterion but a direct consequence of thresholding an unnormalised distance, and it is why $\tau$ is calibrated per backbone ($\tau = 0.3$ for Cosmos, $\tau = 1.2$ for OmniTokenizer). Once $\tau$ is re-tuned to OmniTokenizer's scale, the method recovers a smooth monotonic rate--distortion curve (Figure~\ref{fig:sweep-omni-davis}), confirming that the mechanism transfers and that only its operating point needs relocating.

\paragraph{Selectivity.}
Scale alone does not account for the residual gap, because it is corrected by recalibration. The second axis is the \emph{shape} of the masking-score distribution rather than its position. Measuring the coefficient of variation (CoV, the ratio of standard deviation to mean) of the masking scores, OmniTokenizer is the lower of the two at $0.397$ against Cosmos's $0.569$. A lower CoV means the scores are more tightly concentrated around their mean: relative to its own typical value, OmniTokenizer's latent space draws a weaker distinction between a position that is temporally redundant and one that carries genuine new information. Any single scalar threshold consequently separates the two populations less cleanly, and the mask incurs more errors of both kinds at a matched keep rate. This is an intrinsic property of the representation the backbone has learned, and it bounds how well \emph{any} threshold-based criterion can perform on it, ours included.

\paragraph{Implication.}
Taken together, these two measurements scope the claim made in the main paper. The criterion transfers across continuous backbones, but the quality of the resulting adaptive allocation depends on how well-separated the backbone's temporal-difference statistics are. Backbones whose latent spaces exhibit high masking-score CoV are the more favourable substrate for this family of methods, and CoV is cheap to compute from encoder statistics alone, making it a practical diagnostic for deciding whether a given backbone is a good candidate before any training is undertaken. The same statistics also shorten the calibration itself. In both backbones the threshold we selected independently by visual sweep lands close to the mean of that backbone's masking-score distribution ($\tau = 0.3$ against $\mu = 0.362$ for Cosmos, $\tau = 1.2$ against $\mu = 1.18$ for OmniTokenizer), which is what one would expect of a threshold that partitions positions into roughly the redundant and the informative halves. Setting $\tau \approx \mu$ therefore supplies a principled starting point on a new backbone, obtained from encoder statistics alone, from which a short sweep need only refine.

\phantomsection
\section*{Appendix F: Ablation Analysis and Fill Strategy}
\label{sec:appendix-f}

Table~3 of the main paper varies each of the three design choices at a matched $62\%$ keep rate. Three points are worth expanding. The pixel-space criterion is weakest even though its differences are pooled onto the tokeniser's exact grid in space and in time, so that it ranks the same token slots the latent criterion ranks: a grid-aligned pixel difference still does not track change in the representation actually being compressed. Within the latent space the choice matters far less: L2 is essentially tied with L1, marginally better on FVD and marginally worse on PSNR, SSIM and LPIPS, whereas cosine distance is consistently worse, as it discards the magnitude to which the decoder remains sensitive. And the last-kept reference guards against drift: under a previous-frame reference a change spread slowly over many frames is dropped at every individual step, since no single step exceeds $\tau$, and the deviation from the last transmitted latent accumulates without bound.

\paragraph{The pixel-space criterion in detail.}
Because the pixel row of Table~3(a) is the only ablation whose signal leaves the latent space, we specify it precisely. It is the same pipeline as the latent method with a single substitution, the quantity on which positions are ranked; the thresholding rule, the reference scheme, the fill applied to dropped positions and the frozen encoder and decoder are all unchanged. The comparison therefore isolates the ranking signal and nothing else.

Ranking on pixels while dropping latent tokens requires the two to be placed in correspondence, so the RGB volume is pooled onto the tokeniser's own grid, yielding one value per token. Spatially this is an average over each $8 \times 8$ block, matching the backbone's spatial compression. Temporally we reproduce the backbone's frame grouping exactly rather than approximating it with uniform bins: writing $x_j$ for input frame $j$, $\bar{x}_k$ for pooled latent frame $k$, and $g = (T-1)/(t-1)$ for the number of input frames the encoder folds into each latent frame,
\begin{equation}
    \bar{x}_0 = x_0, \qquad
    \bar{x}_k = \frac{1}{g} \sum_{j = g(k-1)+1}^{gk} x_j \quad (k \geq 1).
    \label{eq:causal-pool}
\end{equation}
For our clip configuration ($T = 33$, $t = 9$) this gives $g = 4$: the first latent frame sees input frame $0$ alone, and every subsequent one averages precisely the four input frames the encoder compresses into it. The pooled grid is then in one-to-one correspondence with the latent grid, and a keep decision taken on a pooled cell is a keep decision on the latent token at the same index. Selection proceeds exactly as in Algorithm~\ref{alg:temporal-l1} of the main paper, with the last-kept reference carried per position on the pooled grid: a cell is dropped when its channel-mean absolute difference from that reference falls below the threshold, and the reference advances only where a cell is kept.

\paragraph{Matching the keep rate.}
Pixel and latent differences are not on a common scale, so the two criteria cannot share a threshold. We recalibrate by bisecting on the pixel threshold until the emergent keep rate meets the latent method's $62\%$ on DAVIS, which yields $\tau_{\text{pixel}} = 0.042$ against $\tau = 0.3$ in latent space. Both rows thus transmit the same number of tokens through the same fill and decoder, and differ only in which tokens those are. That the pixel criterion still loses $1.95$~dB under these conditions is the substance of the claim: the deficit is in the choice of positions, not in the budget or the reconstruction.

\begin{figure}[H]
    \centering
    \includegraphics[width=\linewidth]{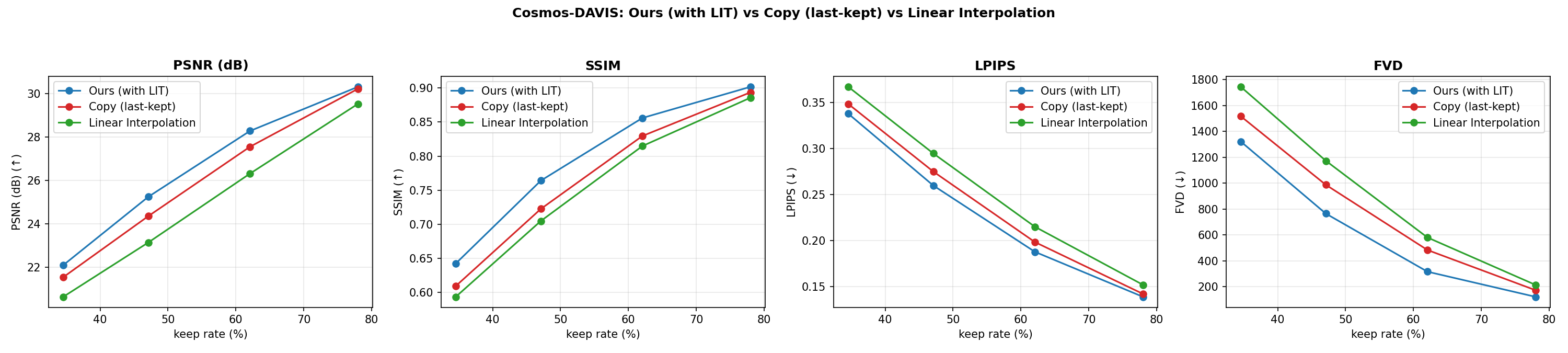}
    \caption{
        Fill strategy across operating points on Cosmos--DAVIS: LIT against a zero-order hold
        (copy last-kept) and linear interpolation between retained latents. LIT leads on all four
        metrics at every keep rate, with the margin widening as compression becomes more aggressive.
    }
    \label{fig:fill-sweep}
\end{figure}

\paragraph{Fill strategy across operating points.}
The fill ablation in the main paper is likewise reported at a single matched keep rate of $62\%$. Since that is one point on a curve, we repeat the comparison across the operating range to confirm the ordering is not an artefact of the chosen budget. Figure~\ref{fig:fill-sweep} compares LIT against the two training-free fills, copying the last-kept latent forward (a zero-order hold) and linearly interpolating between retained latents, on Cosmos--DAVIS at keep rates from $35\%$ to $78\%$.

LIT dominates on every metric at every operating point, and the ordering among the three is stable throughout: LIT, then copy, then interpolation. Two features are worth noting. The margin \emph{widens} as the keep rate falls, which is the expected behaviour, since longer drop-runs leave more for the inpainter to recover and less for a hold or an interpolation to bridge. Conversely the three converge towards the high-keep-rate end: by $78\%$ retention, drop-runs are short enough that copying the last-kept latent is nearly as good as reconstructing it, and the learned inpainter earns little. This locates the regime in which LIT is worth its parameters, and it is the aggressive-compression regime the method targets.

\phantomsection
\section*{Appendix G: Temporal Flicker}
\label{sec:appendix-g}

Reconstruction quality is summarised by PSNR, SSIM, LPIPS and FVD. The first three are computed per-frame and averaged, so they are by construction blind to errors caused by fluctuation. FVD, while aware, is merely a distributional score that doesn't reliably localise the effect.

\paragraph{Why masking induces it.}
The criterion retains and drops different spatial positions at different timesteps. A position that is transmitted at $t$ and inpainted at $t+1$ is reconstructed by two different mechanisms on consecutive frames, so its residual error is not constant over time even when its magnitude is small. Perceptually this reads as flicker: regions individually plausible in each frame but unstable across frames.

\paragraph{Measurement.}
Let $e_t = \hat{x_t} - x_t$ be the error field for frame $t$. We warp $e_t$ into frame $t+1$. We use the Optical flow $\mathcal{W}$ and measure
\begin{equation}
    \mathcal{F} = \bigl\lVert e_{t+1} - \mathcal{W}(e_t) \bigr\rVert_1
    \label{eq:flicker-metric}
\end{equation}

over pixels and incorporate a forward-backward flow consistency check. Without this check, this measure would be dominated by scene motion displacing the error field in lieu of genuine temporal instability.

\begin{table}[htbp]
\centering
\small
\setlength{\abovecaptionskip}{8pt}
\begin{tabular}{lcccc}
\toprule
Region & Backbone & With masking & Masking adds & Backbone share \\
\midrule
Static & 1.190 & 2.156 & $+0.966$ \,($+81.1\%$) & 55.2\% \\
Moving & 3.920 & 5.357 & $+1.436$ \,($+36.6\%$) & 73.2\% \\
\midrule
All    & 3.038 & 4.280 & $+1.242$ \,($+40.9\%$) & 71.0\% \\
\bottomrule
\end{tabular}
\caption{Flicker $\mathcal{F}$ decomposed by region on DAVIS ($n{=}30$, keep rate $0.630$).
``Backbone'' is the frozen tokeniser with no positions dropped. Masking nearly doubles
flicker in static regions but adds roughly a third in moving ones, while the backbone remains
the majority contributor throughout.}
\label{tab:flicker-decomposition}
\end{table}

Two observations are apparent. Masking increases flicker relative to the full-rate backbone, but the
backbone is itself far from flicker-free and remains the primary contributor
(Table~\ref{tab:flicker-decomposition}); masking adds to an already-unstable error field rather
than creating instability where there was none. The asymmetry between regions is the more
informative result: our contribution to the artefact is concentrated where it is most visible.

The last-kept reference is also more stable than a previous-frame reference, mirroring the
fidelity ordering of the reference ablation in the main paper and pointing to a shared cause:
under a previous-frame reference, long drop-runs are inpainted against a reference that has
itself drifted, so the correction applied when a position is finally retained arrives as an
abrupt step.

\paragraph{A temporally-regularised objective.}
The objective of the main paper supervises each frame independently, through a pixel term and
a latent term, and therefore provides no gradient penalising temporal inconsistency; the
artefact induced by the criterion goes uncorrected during training. A natural remedy is to
supervise temporal \emph{differences} rather than frames alone, adding a pixel-space
temporal-L1 term that matches consecutive-frame differences between reconstruction and source,
\begin{equation}
    \mathcal{L}_{\text{temporal}} =
    \bigl\lVert (x_t - x_{t-1}) - (\hat{x}_t - \hat{x}_{t-1}) \bigr\rVert_1 ,
    \label{eq:temporal-l1-loss}
\end{equation}
which is the un-compensated analogue of Equation~\ref{eq:flicker-metric}. Expanding the terms
gives $(x_t - x_{t-1}) - (\hat{x}_t - \hat{x}_{t-1}) = e_{t-1} - e_t$: the term penalises
temporal variation of the reconstruction \emph{error}, with motion cancelling exactly. It
therefore suppresses flicker without penalising genuine movement.

\begin{table}[htbp]
\centering
\small
\setlength{\abovecaptionskip}{8pt}
\begin{tabular}{lcccc}
\toprule
 & \multicolumn{2}{c}{Flicker $\mathcal{F}$ $\downarrow$} & \multicolumn{2}{c}{Fidelity} \\
\cmidrule(lr){2-3}\cmidrule(lr){4-5}
Objective & All & Static & Error level $\downarrow$ & PSNR $\uparrow$ \\
\midrule
Frame-wise only (main paper) & 4.280 & 2.156 & 6.986 & 28.27 \\
$+\ \mathcal{L}_{\text{temporal}}$ (pixel) & \textbf{4.249} & \textbf{2.121} & \textbf{6.845} & 28.11 \\
$+\ $ static-weighted variant & 4.251 & 2.123 & -- & -- \\
\bottomrule
\end{tabular}
\caption{Effect of the temporal term on DAVIS ($n{=}30$, keep rate $0.630$). All arms are
taken at the same training step from a shared checkpoint and differ only in objective.}
\label{tab:temporal-objective}
\end{table}

\begin{table}[htbp]
\centering
\small
\setlength{\abovecaptionskip}{8pt}
\begin{tabular}{lccc}
\toprule
Comparison (region) & Mean diff. & 95\% CI & $p$ \\
\midrule
Frame-wise vs.\ $+\mathcal{L}_{\text{temporal}}$ (all)     & $+0.030$ & $[+0.013, +0.048]$ & $0.002$ \\
Frame-wise vs.\ $+\mathcal{L}_{\text{temporal}}$ (static)  & $+0.035$ & $[+0.001, +0.069]$ & $0.011^{\dagger}$ \\
Frame-wise vs.\ $+\mathcal{L}_{\text{temporal}}$ (error level) & $+0.141$ & $[+0.029, +0.252]$ & $0.020$ \\
\midrule
Pixel vs.\ latent form (all)                               & $-0.013$ & $[-0.026, +0.001]$ & $0.070$ \\
Pixel vs.\ static-weighted (static)                        & $-0.002$ & $[-0.004, +0.001]$ & $0.179$ \\
\bottomrule
\end{tabular}
\caption{Paired comparisons over the same 30 clips; positive favours the temporal objective.
Pairing is exact -- every arm is scored on identical clips with identical ground-truth flow --
and is necessary here, as between-clip variance exceeds between-objective variance by roughly
an order of magnitude. $^{\dagger}$Wilcoxon signed-rank; the paired $t$-test gives $p=0.056$.}
\label{tab:temporal-paired}
\end{table}

The pixel-space term reduces flicker significantly and improves error level at the same time,
so the gain is not obtained by blurring (Tables~\ref{tab:temporal-objective}
and~\ref{tab:temporal-paired}). The cost is $0.16$\,dB PSNR and $0.003$ SSIM, with LPIPS and
FVD approximately unchanged. A latent-space analogue performs comparably
($p=0.070$), and weighting the term toward static regions gives no further benefit
($p=0.179$) even when $\lambda$ is rescaled so that both variants contribute equally to the
objective.

Thus, the trade-off has been
characterised at a single operating point, and the majority of the residual flicker originates
in the frozen backbone rather than in the inpainting network, so the achievable gain from any
objective acting on LIT alone is bounded. Establishing the trade-off across operating points,
and whether the artefact can be reduced substantially without fine-tuning the decoder, is left
to future work.

\end{document}